\documentclass[10pt,twocolumn,letterpaper]{article}

\usepackage{iccv}
\usepackage{times}
\usepackage{epsfig}
\usepackage{graphicx}
\usepackage{tabularx}
\usepackage{amsmath}
\usepackage{amssymb}
\usepackage{bm}
\usepackage{bbm}
\usepackage{algorithm}
\usepackage{algorithmic}
\usepackage{booktabs}
\usepackage{multirow}
\usepackage{tabularx}
\usepackage{xcolor}
\usepackage{subfig}
\colorlet{GRAY}{gray}

\usepackage{array}
\newcolumntype{?}{!{\vrule width 1pt}}
\newcolumntype{C}[1]{>{\centering\arraybackslash\hspace{0pt}}p{#1}}

\graphicspath{{images/}}

\usepackage{color}

\newcommand{\comment}[1]{}

\newif\ifdraft
\draftfalse

\ifdraft
 \newcommand{\bt}[1]{{\color{blue}{#1}}}
 \newcommand{\BT}[1]{{\color{blue}{\bf #1}}}
 \newcommand{\pf}[1]{{\color{red}{#1}}}
 \newcommand{\PF}[1]{{\color{red}{\bf #1}}}
 \newcommand{\PMN}[1]{{\color{green}{\bf #1}}}
 \newcommand{\pmn}[1]{{\color{green}{#1}}}
 \newcommand{\MS}[1]{{\color{magenta}{\bf #1}}}
 \newcommand{\ms}[1]{{\color{magenta}{#1}}}
\else
 \newcommand{\bt}[1]{#1}
 \newcommand{\BT}[1]{}
 \newcommand{\pf}[1]{#1}
 \newcommand{\PF}[1]{}
 \newcommand{\pmn}[1]{#1}
 \newcommand{\PMN}[1]{}
 \newcommand{\ms}[1]{#1}
 \newcommand{\MS}[1]{}
\fi

\newcommand{\bx}[0]{\mathbf{x}}
\newcommand{\by}[0]{\mathbf{y}}

\newcommand{\bX}{\mathbf{X}}

\newcommand{\bI}{\mathbf{I}}

\newcommand{\bZ}{\mathbf{Z}}

\newcommand{\bw}{\mathbf{w}}
\newcommand{\bi}{\mathbf{i}}

\newcommand{\real}{\mathbb{R}}
\DeclareMathOperator*{\concat}{\textrm{concat}}

\newcolumntype{M}[1]{>{\centering\arraybackslash}m{#1}}
\newcolumntype{R}[1]{>{\raggedleft\arraybackslash}m{#1}}
\newcolumntype{P}[1]{>{\centering\arraybackslash}p{#1}}

\definecolor{mygray}{gray}{0.2}

\usepackage[pagebackref=true,breaklinks=true,letterpaper=true,colorlinks,bookmarks=false]{hyperref}

\iccvfinalcopy
\def\iccvPaperID{1617} 

\DeclareMathOperator*{\argmin}{\arg\!\min}

\ificcvfinal\fi
\begin{document}

\title{Learning to Fuse 2D and 3D Image Cues for Monocular Body Pose Estimation}

\author{Bugra Tekin \quad\quad Pablo M\'arquez-Neila  \quad\quad Mathieu Salzmann \quad\quad Pascal Fua \\
	EPFL, Switzerland\\
	{\tt\small \{bugra.tekin,pablo.marquezneila,mathieu.salzmann,pascal.fua\}@epfl.ch}
}

\maketitle

\begin{abstract}

Most  recent approaches  to  monocular 3D  human pose  estimation  rely on  Deep Learning. They typically  involve regressing from an image to either 3D joint coordinates directly or 2D joint locations from which 3D coordinates are inferred. Both approaches have their strengths and weaknesses and we therefore propose a novel architecture designed to deliver the best of both worlds by performing both simultaneously and fusing the information along the way. At the heart of our framework is a trainable fusion scheme that learns how to fuse the information optimally instead of being hand-designed. This yields significant improvements upon the state-of-the-art on standard 3D human pose estimation benchmarks.

\end{abstract}


\section{Introduction}

Monocular 3D human pose estimation is a longstanding problem of Computer Vision. Over the years, 
two main classes of approaches have been proposed: Discriminative ones that directly regress 
3D pose from image data~\cite{Agarwal04a,Bo10,Kanaujia07,Pavlakos16,Rosales02,Urtasun08} and 
generative ones that search the pose space for a plausible body configuration that aligns with
the image data~\cite{Gall10,Sidenbladh00,Urtasun06a}. With the advent of ever larger datasets~\cite{Ionescu14a},
models have evolved towards deep architectures, but the story remains largely
unchanged. The state-of-the-art approaches can be roughly grouped into those that directly regress 
3D pose from images~\cite{Ionescu14a,Li14a,Tekin16b,Tekin16a} and those that first predict a 2D pose 
in the form of joint location confidence maps and fit a 3D model to this 2D prediction~\cite{Bogo16,Zhou16a}.



\begin{figure}[t]
	\centering
	\begin{tabular}{c}
		\includegraphics[width=\linewidth]{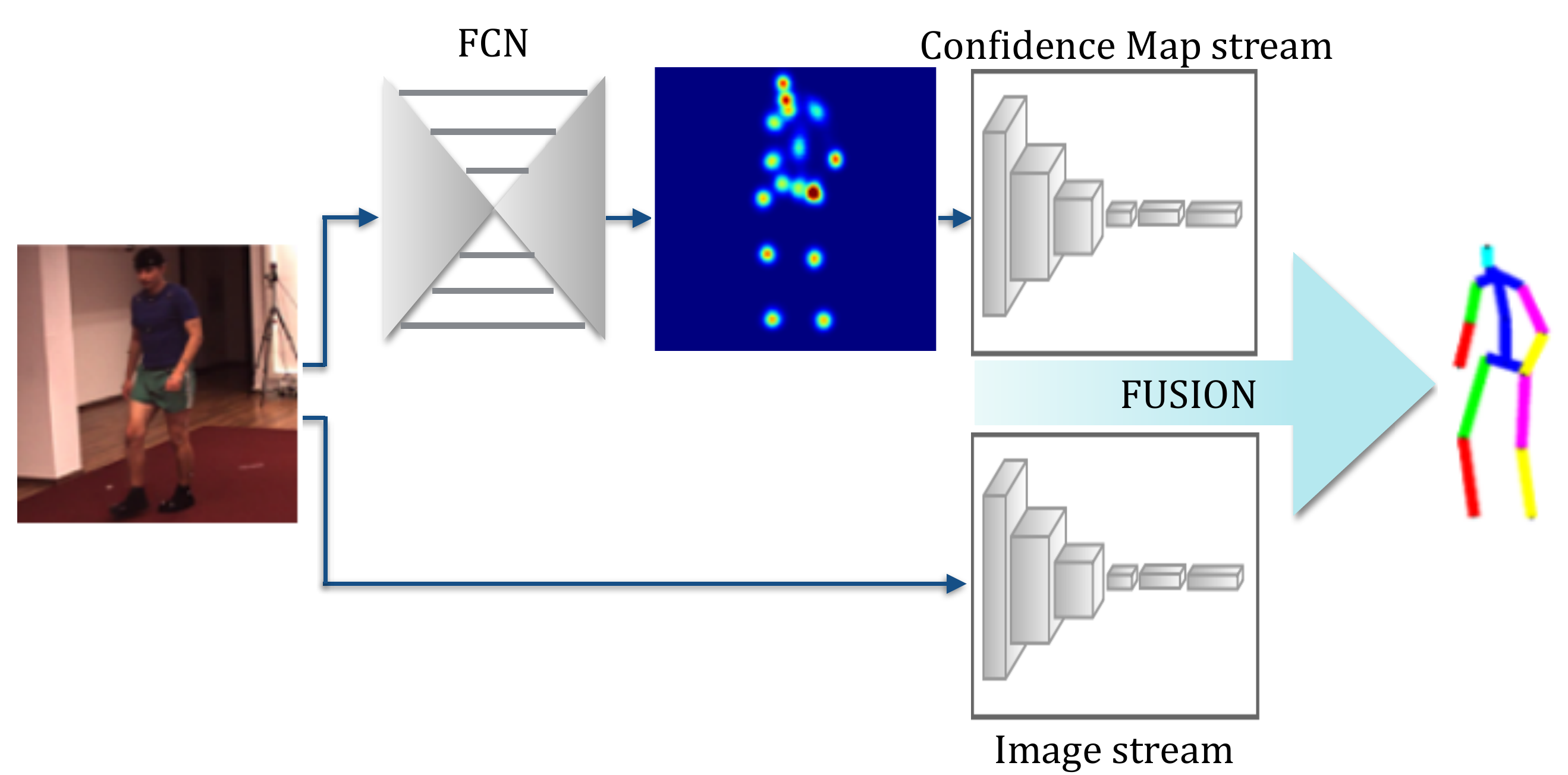}
	\end{tabular}
	\caption{{\bf Overview of our approach.} One stream of our network accounts for the
		 2D joint locations and the corresponding uncertainties. The second one leverages 
		 all 3D image cues by directly acting on the image. The outputs of these two streams 
		 are then fused to obtain the final 3D human pose estimate.}
	\label{fig:intro}
\end{figure}

Since detecting the 2D image location of joints in easier than directly inferring the 3D pose, 
it can be done more reliably. However,  inferring a 3D pose from these 2D locations is fraught 
with ambiguities  and the above-mentioned methods usually rely on a database of 3D models to 
resolve them, at the cost of a potentially expensive run-time fitting procedure. By contrast, 
the methods that regress directly to 3D avoid this extra step but also do not benefit of the 
well-posedness of the 2D joint detection location problem. 

In this paper, we propose the novel architecture depicted by Fig.~\ref{fig:intro} designed 
to deliver the best of both worlds. The first stream, which we will refer to as the {\it Confidence Map Stream}, 
first computes a heatmap of 2D joint locations and then infer the 3D poses from it. The second stream, 
which we will dub the {\it Image Stream}, is designed to produce features that complement those computed by the 
first stream and can be used  in conjunction with them to compute the 3D pose, that is, guide the regression 
process given the 2D locations. 

However, for this approach to be beneficial, effective fusion of the two streams is crucial. In theory, it could happen at any stage of the two streams, ranging from early to late fusion, with no principled way to choose one against the other. We therefore also developed a \emph{trainable fusion} scheme that learns how to fuse the two streams. 


\comment{This is achieved by introducing an additional fusion 
stream acting on the concatenation of the outputs of the confidence map and image streams. We then make use of a trainable 
weight vector to determine at what stage the overall model should switch from the 
two individual confidence map and image streams to the fusion one.}

Ultimately, our approach allows the network to still exploit image cues while inferring 3D poses from 2D joint locations. As we demonstrate in our experiments, the features computed by both streams are decorrelated  and therefore truly encode complementary information. Our contributions can be summarized as follows:
\vspace{-1mm}
\begin{itemize}

\item We introduce a discriminative fusion framework to simultaneously exploit 2D joint location confidence maps and 3D image 
cues for 3D human pose estimation.

\vspace{-1mm}
\item We introduce a novel trainable fusion scheme, which automatically learns where and how to fuse these two sources of information.
\vspace{-1mm}
\end{itemize}
We show that our approach significantly outperforms the state-of-the-art results  on standard benchmarks and yields  accurate pose estimates from images acquired in unconstrained outdoors environments.

\vspace{-1mm}
\section{Related Work}

\vspace{-2mm}
\comment{Over the years,  monocular 3D human pose estimation has  received much attention
in Computer  Vision. } 

The  existing 3D human pose estimation approaches can  be roughly  categorized into
discriminative and  generative ones. In what follows,  we review both types  of approaches.
\comment{with a particular focus on the state-of-the-art}

Discriminative methods aim  at predicting 3D pose directly from  the input data,
may                      it                       be                      single
images~\cite{Ionescu14b,Ionescu11,Kostrikov14,Li14a,Li15a,Pavlakos16,Rogez12,Rosales00,Tekin16b,Yu13},       depth
images~\cite{Girshick11,PonsMoll15,Shotton12},  or   short  image   sequences~\cite{Tekin16a}.   Early
approaches   falling  into   this  category   typically  worked   by  extracting
hand-crafted  features  and  learning  a  mapping  from  these  features  to  3D
poses~\cite{Agarwal04a,Bo10,Ionescu14b,Ionescu11,Kostrikov14,Rosales02,Urtasun08}.
Unsurprisingly,   the   more    recent   methods   tend   to    rely   on   Deep
Networks~\cite{Li14a,Tekin16b,Tekin16a, Zhou16b}.   In  particular,~\cite{Li14a,Tekin16a}
rely on  2D poses to  pretrain the  network, thus exploiting  the commonalities
between  2D and  3D  pose  estimation. In  fact,~\cite{Li14a}  even proposes  to
jointly predict 2D and 3D poses. However, in such approaches, the two predictions are not coupled.
By contrast,~\cite{Park16} introduces a network that uses 2D information for 3D pose estimation. 
	This method, however, does not exploit pixelwise joint location uncertainty, and only makes use of 
	the 2D evidence late in the pose estimation process.
While  these methods exploit the available  3D image cues, they fail to explicitly model 2D joint location 
uncertainty, which matters when addressing a problem as ambiguous as monocular 3D pose estimation. \comment{which, as we demonstrate in our experiments, is crucial when resolving ambiguities of 
3D human pose. \MS{Do we really demonstrate this explicitly?}}

\comment{matters  when addressing a problem as ambiguous as monocular 3D pose estimation.}

Since pose estimation  is much better-posed in  2D than in 3D, a  popular way to
infer joint positions  is to  use  a  generative 
model  to  find  a 3D  pose  whose projection aligns  with the 2D image  data.  
In the past,  this usually involved inferring a 3D human pose by optimizing  an  energy   function  derived  from  
image   information,  such  as silhouettes~\cite{Balan07a,Chen10,Gall10,Gammeter08,Guan10a,Jain10,Ormoneit00,PonsMoll11,Sidenbladh00},
trajectories~\cite{Zhou14a}, feature descriptors~\cite{Sanzari16,Simo-Serra13,Simo-Serra12}
and 2D joint locations~\cite{Akhter15,Amin13,Andriluka10,Fan14b,Kirk05,Ramakrishna12a,Salzmann10a,Urtasun06a,Valmadre10}. 
Another class of approaches retrieve the pose from a dictionary
of 3D poses based on similarity with the 2D image evidence~\cite{Efros03,Howe11,Li15a,Mori02,Mori06}.
With the growing availability of large  datasets and  the advent  of Deep Learning,  the emphasis  has shifted
towards using discriminative 2D pose
regressors~\cite{Carreira16,Chen14,Chu16,Du12,Gkioxari16,Insafutdinov16a,Jain14,Newell16,Pfister15,Pishchulin16a,Toshev14,Wei16,Yang11}
to extract the 2D pose and infer a 3D one from it~\cite{Bogo16,Elhayek15,Yasin16,Zhou16a}.
The  2D joint locations are  represented  by  heatmaps that  encode  the confidence  of
observing  a  particular  joint  at  any given  image  location.   A  human  body
representation,  such   as  a   skeleton~\cite{Zhou16a},  or  a   more  detailed
model~\cite{Bogo16} can then  be fitted to these predictions.   While this takes
2D joint positions into  account, it  ignores  image  information during  the  fitting
process. It  therefore discards  potentially important 3D  cues that  could help
resolve ambiguities.

\section{Approach}
\label{sec:approach}

Our goal is to increase the robustness and accuracy of monocular 3D pose estimation by exploiting 
image cues to the full  while also taking advantage of the fact that 2D joint locations can be 
reliably detected by modern CNN architectures. To this end, we designed the two stream architecture 
depicted by Fig.~\ref{fig:intro}. The Confidence Map Stream shown at the top first computes a heatmap
of 2D joint locations from which feature maps can be computed. The Image Stream shown at the bottom 
extracts additional features directly from the image and all these features are fused to produce a final 3D pose vector. 


\begin{figure}
	\def\tabularxcolumn#1{m{#1}}
	\begin{tabularx}{\linewidth}{@{}cc@{}}
		\begin{tabular}{ccc}
			\subfloat[Early fusion]{\includegraphics[width=\linewidth]{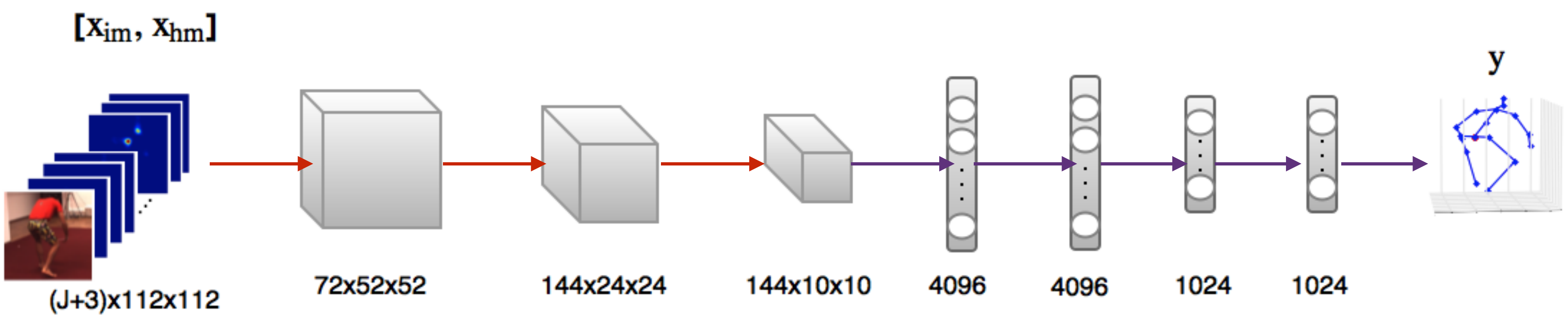}} \\
			\subfloat[Fusion at a specific layer]{\includegraphics[width=\linewidth]{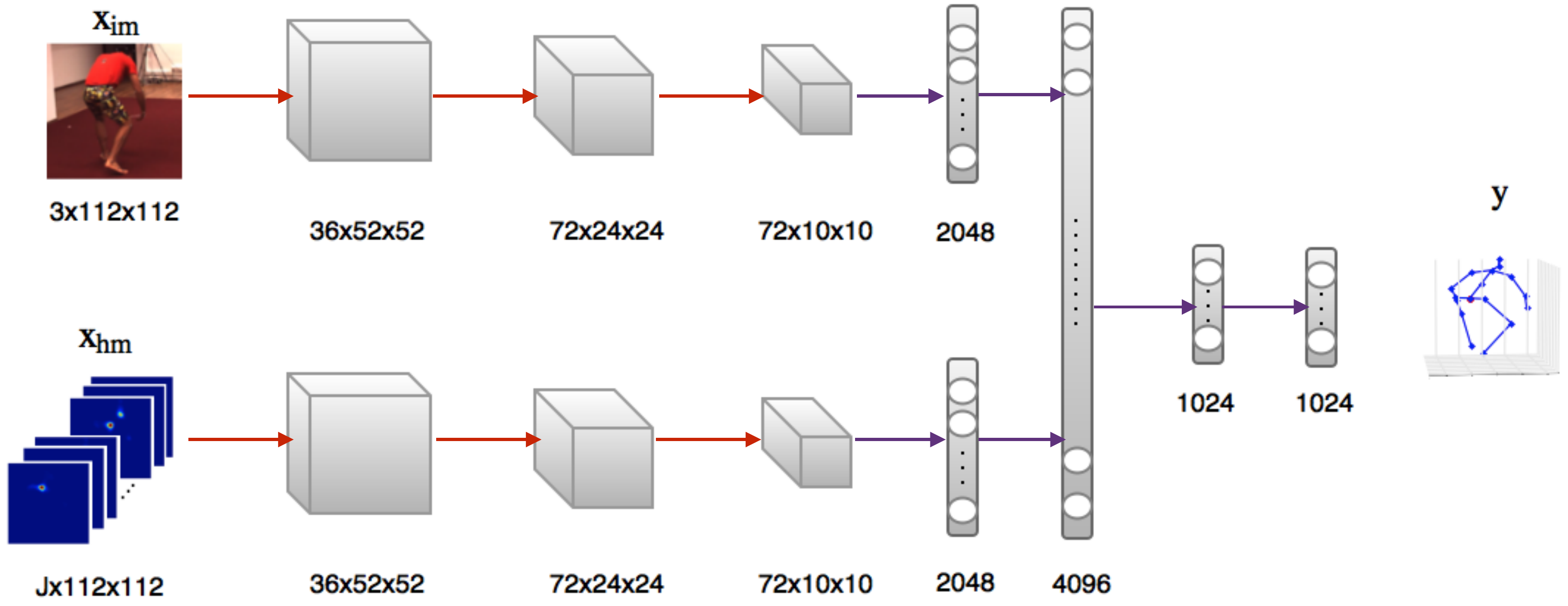}} \\
			\subfloat[Late fusion]{\includegraphics[width=\linewidth]{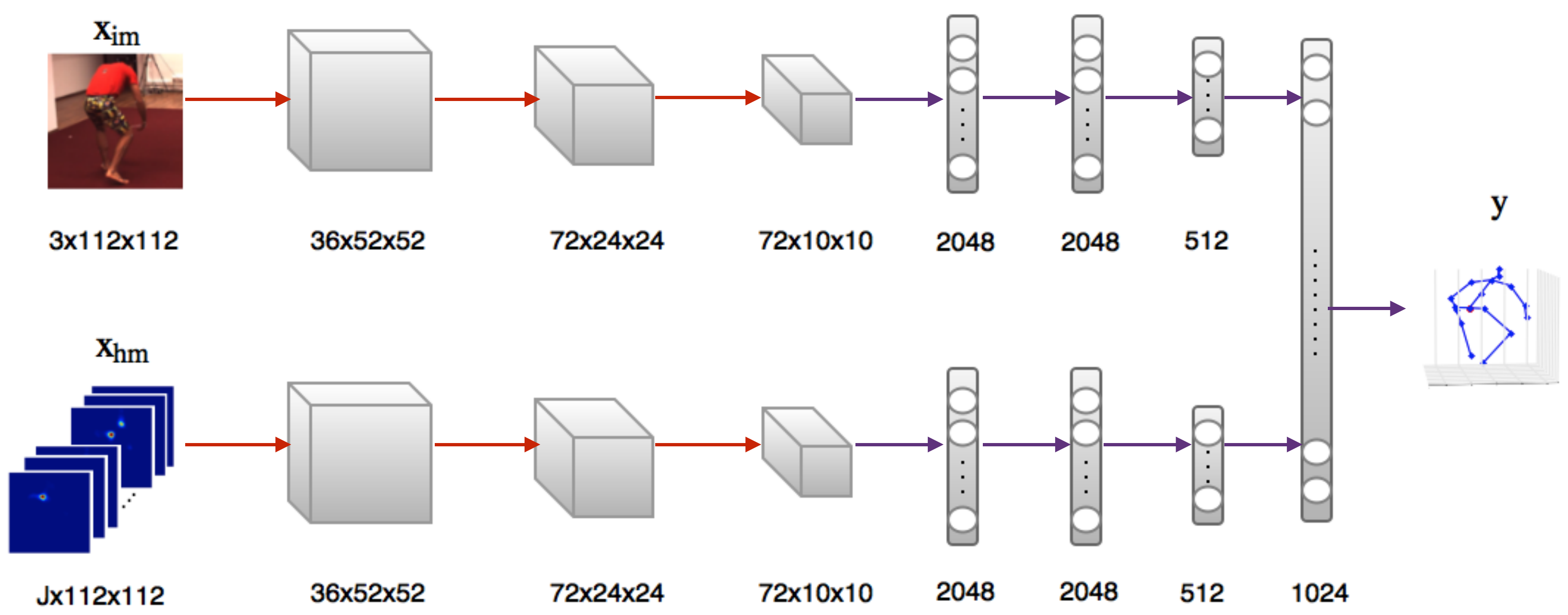}} \\
		\end{tabular}

	\end{tabularx}
	
	\caption{{\bf Three different instances of hard-coded fusion.}
	 \pf{ The fusion strategies combine  2D joint location confidence maps  with 3D cues directly  extracted from the
			input  image.}}
	\label{fig:hardcoded}
\end{figure}

\begin{figure*}
	\centering
	\scalebox{0.94}{
	\begin{tabular}{c}
	\includegraphics[width=0.7\linewidth, height=8cm]{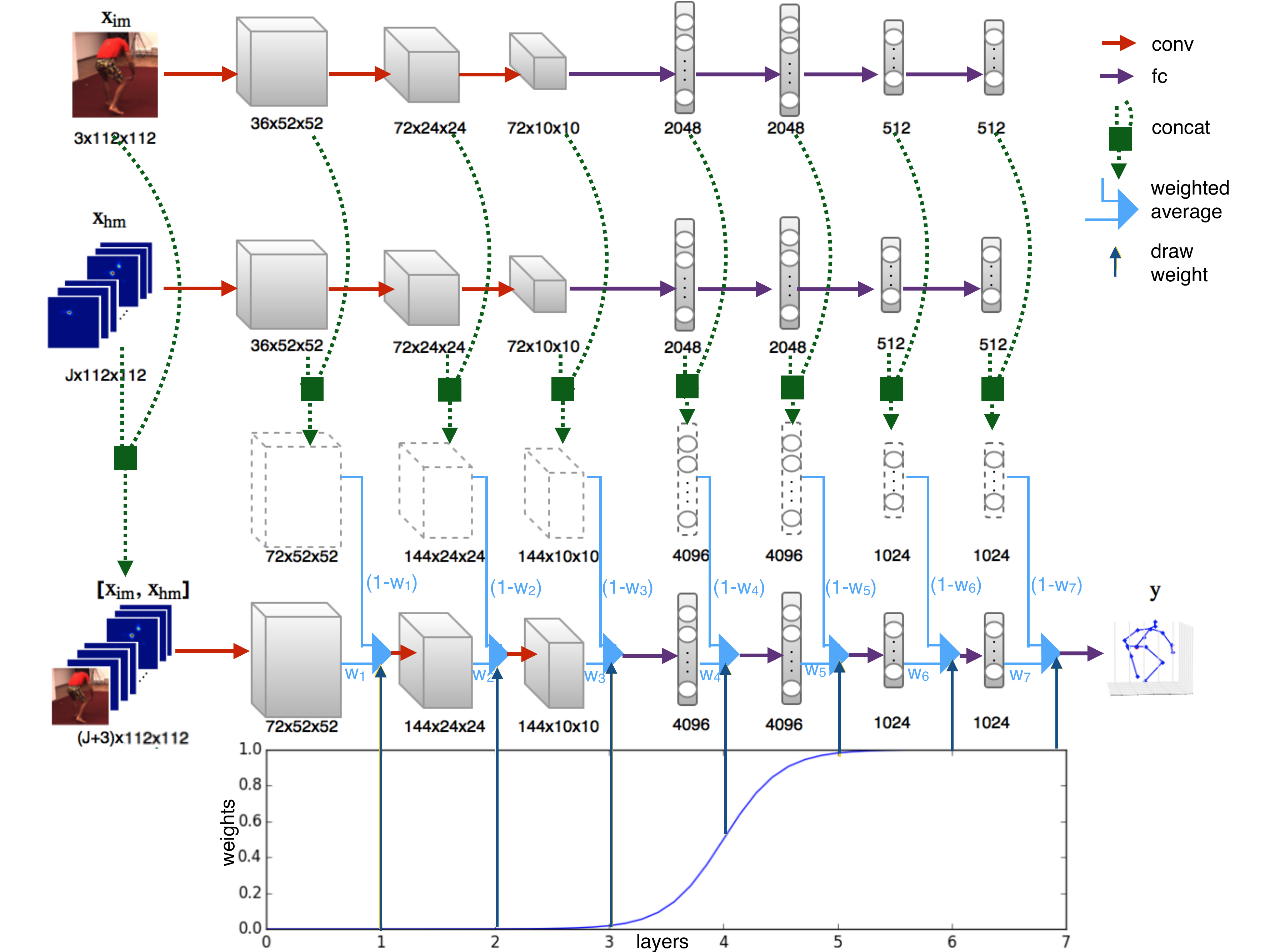}
	\end{tabular}
}
	\caption{{\bf Trainable fusion architecture.} The first two streams take as input the image
		and 2D joint location confidence maps, respectively. The combined feature maps of 
		the image and confidence map stream are fed into the fusion stream and linearly combined
		with the outputs of the previous fusion layer. The linear combination of the streams is controlled
		by a weight vector shown at the bottom part of the figure.  The numbers
		below each layer  represent the corresponding size of  the feature maps
		for convolutional layers and the number of neurons for fully connected
		ones.}
	\label{fig:trainable}
\end{figure*}

As shown in Fig.~\ref{fig:hardcoded}, there is a whole range of ways to perform the fusion of these two 
data streams, ranging from early to late fusion with no obvious way to choose the best, which might well 
be problem-dependent anyway. To solve this conundrum, we rely on the fusion architecture depicted by 
Fig.~\ref{fig:trainable}, which involves indroducing a third {\it fusion stream} that combines the 
feature maps produced by the two data streams in a trainable way. Each layer of the fusion stream  
acts on a linear combination of the previous fusion layer with the concatenation of the two data 
stream outputs.  In effect, different weight values for these linear combinations correspond to 
different fusion strategies.


In  the  remainder of  this  section, we formalize  this generic architecture and study different ways to set these weights, including learning them along with the weights of the data streams, which is the approach we advocate. 

\subsection{Fusion Network}

	Let $\{\bI_l\}_{l=0}^L$ be the feature maps of the \emph{image stream} and
	$\{\bX_l\}_{l=0}^L$ be the feature maps of the \emph{confidence map stream}.
	As special cases, 
	$\bI_0: [1,3]\times[1,H]\times[1,W] \rightarrow [0,1]$ is the input RGB image,
	and $\bX_0: [1,J]\times[1,H]\times[1,W]  \rightarrow \real_+$ are the confidence maps encoding the probability of observing each
	one  of  $J$~body  joints  at  any given  image location. The feature maps~$\bI_l$ and~$\bX_l$
	at each layer~$l$ must coincide in width and height but can have different number
	of channels. In the following, we denote each feature map at level $l$ as both the output
	of layer~$l$ and the input to layer~$l+1$.
	
	Let $\{\bZ_l\}_{l=0}^{L+1}$ be the feature maps of the \emph{fusion stream}.
	The feature map~$\bZ_l$ is the output of layer~$l$, but, unlike
	in the data streams, the input to layer~$l+1$ is a linear
	combination of~$\bZ_l$ with $\bI_l$ and~$\bX_l$ given by
	{\begin{equation}
	(1-w_l)\cdot\concat(\bI_l, \bX_l) + w_l \cdot\bZ_l,\quad 1\le l \le L,
	\end{equation}
	where $\concat(\cdot,\cdot)$~is the concatenation of the given feature maps along the channel axis, and
	$w_l$ is the $l$-th~element of the fusion weights~$\bw \in [0, 1]^L$ controlling the mixture.
	For this mixture to be possible, $\bZ_l$ must have the same size as $\bI_l$ and~$\bX_l$ and a number of
	channels equal to the sum
	of the number of channels of $\bI_l$ and~$\bX_l$. As special cases, $\bZ_0 = \concat(\bI_0, \bX_0)$, and
	$\bZ_{L+1}\in \mathbb{R}^{3J}$ is the output of the network, that is, the $J$~predicted
	3D joint locations.
	
	In essence, the fusion weights~$\bw$ control where and how the fusion of the data streams occurs. Different
	settings of these weights lead to different fusion strategies. 
	We illustrate this with two special cases below, and then introduce an approach to automatically learn these weights \ms{together with the other network parameters.}

\vspace{-0.4cm}
\paragraph{Early fusion.}
If the fusion weights are all set to one, $\bw = \mathbf{1}$, the two data streams are ignored,
and only the fusion one is considered to compute the output. Since the fusion stream takes
 the concatenation of the image~$\bI_0$ and the confidence maps~$\bX_0$ as input,
this is equivalent to the early fusion architecture of Fig.~\ref{fig:hardcoded}(a).

\vspace{-0.4cm}
\paragraph{Fusion at a specific layer.}
Instead of fusing the streams in the very first layer, one might want to postpone the fusion point to a later
layer~$\beta \in \{0,\cdots,L\}$. 
In our formalism, this can be achieved by setting the
fusion weights to~$w_l = \mathbb{I}[l > \beta]$, where $\mathbb{I}$~is the indicator function.
For example, when $\beta=4$, our network
becomes equivalent to the one depicted by Fig.~\ref{fig:hardcoded}(b). The early
 and late fusion architectures of Fig.~\ref{fig:hardcoded}(a, c) can also be represented in this manner by setting~$\beta=0$ and~$\beta=L$, respectively.

	\vspace{0.2cm}
	Ultimately, the complete fusion network encodes a function~$f(\bi, \bx; \theta, \bw)=\left.\bZ_{L+1}\right|_{\bI_0=\bi, \bX_0=\bx}$
	mapping from an image~$\bi$ and confidence maps~$\bx$ to the 3D joint locations,
	parametrized by layer weights~$\theta$ and fusion weights~$\bw$.
	With manually-defined fusion weights, given a set of $N$~training pairs $(\bi_n, \bx_n)$ with
	 corresponding ground-truth joint positions~$\by_n$,
	the parameters  $\theta$ can be learnt by minimizing the square loss
	expressed as
	\begin{equation}
	L(\theta) = \sum_{n=1}^N \left\| f(\bi_n,\bX_n; \theta, \bw) - \by_n \right\|_2^2\;.
	\label{eq:training_nofusion}
	\end{equation}

\paragraph{Trainable fusion.}

Setting the weights manually, which in our formalism boils down to choosing $\beta$, is not obvious; the best value for $\beta$ will typically depend on the network architecture, the problem and the nature of the input data. A straightforward approach would consist of training networks for all possible values of $\beta$ to validate the best one, but this quickly becomes impractical.
To address this issue, we introduce a trainable fusion approach, which aims to learn~$\beta$
from data \ms{jointly with the network parameters.}
To this end, however, we cannot directly use the indicator function, which has
zero derivatives almost everywhere, thus making it inapplicable to gradient-based optimization.
Instead, we propose to approximate the indicator function by a sigmoid function
\begin{equation}
    w_l = \dfrac{1}{1 + e^{-\alpha\cdot(l - \beta)}},
    \label{eq:sig}
\end{equation}
parameterized by~$\alpha$ and~$\beta$. As above, $\beta$~determines the stage at which fusion
occurs and $\alpha$~controls how sharp the 
transition between weights with value 0 and with value 1 is. When $\alpha\to\infty$,
the function in Eq.~\ref{eq:sig} becomes equivalent to the indicator function\footnote{Except at $l=\beta$.},
while, when $\alpha=0$, the network mixes the data and
fusion streams in equal proportions at every layer.

\ms{In practice, mixing the data and fusion streams at every layer is not desirable. First, by contrast to having binary weights ${\bf w}$, which deactivate some of the layers of each stream, it corresponds to a model with a very large number of active parameters, and thus prone to overfitting. Furthermore, after training, a model with binary weights can be pruned, by removing the inactive layers in each stream, that is all layers~$l$ from the fusion stream where $w_l \approx 0$, and all layers~$l$ from the data streams where $w_l \approx 1$. This yields a more compact, and thus more efficient network for test-time prediction.}

\ms{To account for this while learning} where to fuse the information sources, we modify the loss function
of Eq.~\eqref{eq:training_nofusion} by incorporating a term that penalizes small values of~$\alpha$ and favors sharp fusions.
This yields a loss of the form
\begin{equation}
    L(\theta, \alpha, \beta) = \sum_{n=1}^N \left\| f(\bi_n,\bX_n; \theta, \alpha, \beta) - \by_n \right\|_2^2 + \dfrac{\lambda}{\alpha^2}\;,
    \label{eq:training_fusion}
\end{equation}
with $\alpha$ and $\beta$ as trainable parameters, in addition to $\theta$, and a 
hyperparameter~$\lambda$ weighing the penalty term.
Altogether, this loss lets us simultaneously
find the most suitable fusion layer~$\beta$ for the given data and the corresponding network parameters 
$\theta$, while encouraging a sharp fusion function to mimic the behavior of the indicator function.

\comment{\PMN{Several readers asked why having a sharp function is desirable. We should provide an argument for this.}
\MS{One might ask why it is desirable to have a sharp fusion function. Do we have an answer to this?}
\PMN{Our contribution is a fusion network. 
A smooth sharp function averages the streams, but it is not fusion (understanding fusion
as several streams being merged into a single one). A reviewer could argue that averaging
streams might be a better idea than fusing them, and they might be right. We don't know.
Maybe we should run an experiment training each~$w_l$ independently. If the result is worse than
fusion (hopefully), we can include it in the supplementary material.}
}
In practice, we initialize $\alpha$ with a small value of $0.1$ and $\beta$ to the middle layer of the complete network.
We use the ADAM~\cite{Kingma15} gradient update method with a learning
rate of~$10^{-3}$ to guide the optimization. We set the regularization parameter to~$5\cdot10^3$,
which renders the magnitude of both the regularization term and the main cost comparable. 
We use dropout and data augmentation to prevent overfitting. 

\comment{Note that the architecture of Fig.~\ref{fig:trainable} is used only for
training purposes. Once training is over and a sharp fusion function has been obtained,
we remove all layers~$l$ from the fusion stream where $w_l \approx 0$,
and all layers~$l$ from the data streams where $w_l \approx 1$, \bt{thus making
test-time prediction more efficient}.}


\subsection{2D Joint Location Confidence Map Prediction}
\label{sec:heatmap}

\comment{
\MS{Do we really want to discuss U-Net in here? In the end, we use the hourglass network, so why should people care about U-Net?}
\BT{Removed the discussion of U-Net.}
}

Our approach depends on generating heatmaps of the 2D joint locations
that we can feed as input to the confidence map stream. To do so, we rely 
on a fully-convolutional network with skip connections~\cite{Newell16}. 
Given an RGB image as  input,  it  performs a  series  
of  convolutions  and  pooling operations  to reduce  its  spatial resolution,  
followed  by upconvolutions  to produce pixel-wise confidence values for each pixel.  
We employed the stacked hourglass network design of~\cite{Newell16}, which carries out
repeated bottom-up, top-down processing to capture spatial relationships in the image.
We perform heatmap regression to assign high confidence values to the most likely 
joint positions. In our experiments, we fine-tuned the hourglass network initially
trained on the MPII dataset~\cite{Andriluka14} using the training data specific to each experiment
as a preliminary step to training our fusion network. In practice, we have observed that \ms{ using the more accurate 
2D joint locations predicted by the stacked network architecture improves the overall 3D prediction accuracy
over using those predicted by a single-stage fully-convolutional network, such as~\cite{Ronneberger15}. Ultimately, these predictions 
provide reliable intermediate features for the 3D pose estimation task.}

\section{Results}
\label{sec:results}

In this section, we first describe the datasets we tested our approach on and the corresponding 
evaluation protocols. We then compare our approach against the state-of-the-art methods and provide 
a detailed analysis of our general framework.

\subsection{Datasets}
\label{ssec:datasets}

We evaluate our  approach on the Human3.6m~\cite{Ionescu14a}, HumanEva-I~\cite{Sigal06}, KTH Multiview
Football II~\cite{Burenius13} and Leeds Sports Pose (LSP)~\cite{Johnson10b} datasets described below.

\noindent{\bf Human3.6m} is a large and diverse motion capture dataset including  3.6 
million  images  with  their corresponding 2D  and 3D  poses. The  poses are viewed  from 4  different 
camera angles. The  subjects carry out  complex motions corresponding to  daily human activities.
 We  use the standard $17$ joint skeleton from Human3.6m as our pose representation.

\noindent{\bf HumanEva-I} comprises synchronized images and motion capture data and is a  standard 
benchmark for 3D human pose  estimation. The output pose is a vector of $15$ 3D joint coordinates.

\noindent{\bf KTH Multiview Football II} provides a benchmark to evaluate the performance of pose 
estimation algorithms in unconstrained outdoor settings. The camera follows a soccer player moving 
around the pitch. The videos are captured from 3 different camera viewpoints. The output pose is a vector of 14 3D joint coordinates.

\noindent{\bf LSP} is a standard benchmark for 2D human pose estimation and does
not contain any ground-truth 3D pose data. The images are captured in unconstrained
outdoor settings. 2D pose is represented in terms of a vector of $14$ joint coordinates.
We report qualitative 3D pose estimation results on this dataset.

\begin{table*}[tbph]
	\begin{center}
		\tabcolsep=0.17cm
		\scalebox{0.73}{
			\begin{tabular}[b]{llccccccccc}
				\toprule
				Input &Method							 &  Directions    &  Discussion     &  Eating        &  Greeting      &  Phone Talk   		  &  Posing          &  Buying        &  Sitting               & Sitting Down    \\
				\toprule
				\multirow{6}{*}{Single-Image}
				&Ionescu et al.~\cite{Ionescu14a}		 &  132.71        & 183.55          & 132.37         & 164.39 	      & 162.12	      		  & 150.61	         & 171.31         & 151.57                 & 243.03\\
				&Li \& Chan~\cite{Li14a}			     &  -	 	      & 148.79          & 104.01         & 127.17		  & -			  		  & -		         & -              & -                      & -	  	\\
				&Li et al.~\cite{Li15a}					 & -              & 134.13          & 97.37          & 122.33         & -             		  & -                & -              & -			           & -       \\
				&Li et al.~\cite{Li16b}					 & -              & 133.51          & 97.60          & 120.41         & -             		  & -                & -              & -			           & -        \\
				&Zhou et al.~\cite{Zhou16a}        		 &-               & -               & -              &-               &-              		  & -                &-               &- 			           & -         \\
				&Rogez \& Schmid~\cite{Rogez16}     	 & -              & -               & -              & -              & -             		  & -                & -              & -			           & - 		    \\
				&Tekin et al.~\cite{Tekin16b}			 & -              & 129.06          & 91.43          & 121.68         & -             		  & -             	 & -           	  & - 					   & -           \\
				&Park et al.~\cite{Park16}			     & 100.34         & 116.19          & 89.96          & 116.49         & 115.34       		  & 117.57       	 & 106.94         & 137.21    			   & 190.82      \\
				&Zhou et al.~\cite{Zhou16b}			     & 91.83          & 102.41          & 96.95          & 98.75          & 113.35       		  & 90.04       	 & 93.84          & 132.16    			   & 158.97      \\
				\midrule
				\multirow{3}{*}{Video}
				&Tekin et al.~\cite{Tekin16a}            & 102.41 	      & 147.72 	        & 88.83  	     & 125.28  	      & 118.02   	          &112.3   	  		 & 129.17         & 138.89    			   &224.90		\\
				&Zhou et al.~\cite{Zhou16a}			     & 87.36          & 109.31          & 87.05          & 103.16         & 116.18		          & 106.88       	 & 99.78          & 124.52    			   & 199.23      \\
				&Du et al.~\cite{Du16}                   & 85.07      	  & 112.68          & 104.90         & 122.05         & 139.08                & 105.93       	 & 166.16         &	117.49     			   & 226.94       \\
				\midrule
				Single-Image & Ours						 & \textbf{54.23} & \textbf{61.41}  & \textbf{60.17} & \textbf{61.23} &\textbf{79.41}         & \textbf{63.14}   & \textbf{81.63} & \textbf{70.14}         & \textbf{107.31}             \\
				\bottomrule
				\toprule
			\end{tabular}
		}
	\end{center}
	\begin{center}
		\tabcolsep=0.1cm
		\scalebox{0.75}{
			\begin{tabular}[b]{llcccccccc}
				Input &Method:						&  Smoking  	&  Taking Photo  & Waiting     	   & Walking     		   &  Walking Dog    &  Walking Pair     	 & Avg. (All) & Avg. (6 Actions)        \\
				\toprule
				\multirow{6}{*}{Single-Image}
				&Ionescu et al.~\cite{Ionescu14a}  	& 162.14    	& 205.94     	 &  170.69	 	   & 96.60 				   & 177.13	         &  127.88       		 & 162.14 		   & 159.99					\\
				&Li \& Chan~\cite{Li14a}			& - 		 	& 189.08         & -               & 77.60				   & 146.59		     & -             		 & -      		   & 132.20 				 	\\
				&Li et al.~\cite{Li15a}             & -         	& 166.15         & -               & 68.51       		   & 132.51          & -                 	 & -			   & 120.17        					\\
				&Li et al.~\cite{Li16b}				& -           	& 163.33         & -               & 73.66       		   & 135.15          & -                  	 & -      	       & 121.55					\\
				&Zhou et al.~\cite{Zhou16a}         &-           	&-               &-                &-            		   & -             	 & -                     & 120.99 	       & - 						\\
				&Rogez \& Schmid~\cite{Rogez16}	    & -           	& -              & -               & -           		   & -               & -                     & 121.20    	   & -   					\\
				&Tekin et al.~\cite{Tekin16b}       & -      		& 162.17         & -               & 65.75       		   & 130.53          & -                 	 & -			   & 116.77             				\\
				&Park et al.~\cite{Park16}          & 105.78    	& 149.55         & 125.12          & 62.64       		   & 131.90          & 96.18                 & 117.34   	   & 111.12      			\\
				&Zhou et al.~\cite{Zhou16b}         & 106.91    	& 125.22         & 94.41           & 79.02       		   & 126.04          & 98.96                 & 107.26   	   & 104.73      			\\
				\midrule
				\multirow{3}{*}{Video}
				&Tekin et al.~\cite{Tekin16a}		& 118.42    	& 182.73         & 138.75          & 55.07 			 	   & 126.29          & 65.76			  	 & 124.97		   & 120.99     				\\
				&Zhou et al.~\cite{Zhou16a}         & 107.42    	& 143.32         & 118.09          & 79.39       		   & 114.23          & 97.70                 & 113.01   	   & 106.07      			\\
				&Du et al.~\cite{Du16}              & 120.02     	& 135.91         & 117.65          & 99.26       		   & 137.36       	 & 106.54            	 & 126.47    	   & 118.69 			\\
				\midrule
				Single-Image &Ours 					& \textbf{69.29}& \textbf{78.31} & \textbf{70.27}  & \textbf{51.79}		   & \textbf{74.28}  & \textbf{63.24}		 & \textbf{69.73}  & \textbf{64.53}		\\
				\bottomrule
			\end{tabular}
		}
	\end{center}
	\vspace{-0.3cm}
	\caption{{\bf   Comparison  of   our   approach  with   state-of-the-art
            algorithms on \emph{Human3.6m}.} We  report 3D joint position errors
          in  mm,  computed  as  the  average  Euclidean  distance  between  the
          ground-truth  and  predicted   joint  positions. `-' indicates  that the
          results  were not  reported for  the  respective action  class in  the
          original  paper.  Note  that our  method consistently  outperforms the
          baselines.  }  
	\label{tab:overall}
	\vspace{-0.3cm}
\end{table*}

\subsection{Evaluation Protocol}
\label{ssec:eval}

On Human3.6m, we used the same data partition as in earlier work~\cite{Li14a,Li15a,Li16b,Tekin16a,Zhou16a}
for a fair comparison. The data from 5~subjects (S1, S5, S6, S7, S8) was used for training and
the data from 2 different subjects (S9, S11) was used for testing. We evaluate the accuracy
of 3D human pose estimation in terms of average Euclidean distance between the predicted
and ground-truth 3D joint positions, as in~\cite{Li14a,Li15a,Li16b,Tekin16a,Zhou16a}. Training and testing
were carried out monocularly in all camera views.

In~\cite{Bogo16}, \cite{Pavlakos16}\footnote{While~\cite{Pavlakos16} also reports results without Procrustes analysis, the authors confirmed to us by email that their evaluation assumes the ground-truth depth of the root joint to be known to go from their volumetric representation to 3D pose in metric space. Since this also sets the scale of the skeleton, we believe that a comparison using the full Procrustes transformation for both their approach and ours is the right one to perform here.} ,  and~\cite{Sanzari16}\footnote{This it is not explicitly stated in~\cite{Sanzari16}, but the authors confirmed this to us by email.}  the estimated skeleton was  first aligned to the  ground-truth one  by Procrustes transformation before measuring the joint distances. This is therefore what we also do when comparing against~\cite{Bogo16,Pavlakos16,Sanzari16}.

\comment{Testing  was
carried  out only  in the  frontal camera  ("cam3") from  trial $1$  using the
sequences from S9  and S11.}

On HumanEva-I, following the standard evaluation protocol~\cite{Bogo16,Simo-Serra13,Tekin16a,Yasin16,Zhou16a}, we trained our model on  the training 
sequences of  subjects S1, S2 and  S3  and  evaluated on  the  validation sequences of all subjects.  
We pretrained  our network on Human3.6m  and  used  only  the  first  camera  view  for  further  
training  and validation.

On the KTH  Multiview Football II  dataset, we evaluate  our method on  the sequence
containing Player~2,  as in~\cite{Belagiannis14a,Burenius13,Pavlakos16,Tekin16a}. Following~\cite{Belagiannis14a,Burenius13,Pavlakos16,Tekin16a}, 
the first half of the sequence from camera~1 is used for training and the second half for
testing. To compare our results to those of~\cite{Belagiannis14a,Burenius13,Pavlakos16,Tekin16a},
we report accuracy using the percentage of correctly estimated parts (PCP) score. Since 
the training set   is    quite   small, we propose to   pretrain our network on the 
recent synthetic dataset~\cite{Chen16}, which contains images of sports players
with their corresponding 3D poses. We then fine-tuned it using the training data 
from KTH  Multiview Football II. We report results with and without this pretraining.

\subsection{Comparison to the State-of-the-Art}

We first compare our approach with state-of-the-art baselines on the \emph{Human3.6m~\cite{Ionescu14a}, 
	HumanEva~\cite{Sigal06} and KTH Multiview Football~\cite{Burenius13}} datasets. 
\comment{Here, \emph{Ours} refer to our \ms{trainable} fusion strategy, which, as
shown in Section~\ref{ssec:analysis}, yields the best results among \comment{our 
four} different \bt{fusion} strategies. \MS{Are we still going to show this?}
\BT{There is going to be a comparison against Early and Late-Late fusion.}}

\vspace{-4mm}

\paragraph{Human3.6m.} In Table~\ref{tab:overall},
we  compare the  results of our trainable fusion approach with  those of  the following  state-of-the-art
single image-based  methods: KDE regression  from HOG features  to 3D
poses~\cite{Ionescu14a}, jointly training a 2D  body part detector and a
3D pose  regressor~\cite{Li14a,Park16}, the maximum-margin  structured learning
framework of~\cite{Li15a,Li16b}, the deep structured prediction approach
of~\cite{Tekin16b}, pose regression with kinematic constraints~\cite{Zhou16b},
and  3D  pose  estimation  with  mocap  guided  data
augmentation~\cite{Rogez16}.   For  completeness,  we also  compare  our
approach  to  the  following  methods   that  rely  on  either  multiple
consecutive images or impose temporal consistency: regression from short
image sequences  to 3D poses~\cite{Tekin16a},  fitting a sparse  3D pose
model to  2D  confidence map predictions across  frames~\cite{Zhou16a}, and
fitting a  3D pose  sequence to  the 2D joints  predicted by  images and
height-maps that encode the height of each pixel in the image with respect 
to a reference plane~\cite{Du16}. 
	
As can be seen from the results in Table~\ref{tab:overall}, our approach
outperforms all the methods on all the action categories
by a large margin. In  particular,  we  outperform   the  image-based  
regression  methods of~\cite{Ionescu14a,Li14a,Li15a,Li16b,Tekin16b,Park16,Zhou16b},   as    
well   as   the model-fitting strategy of~\cite{Li15a,Li16b}.  This, we 
believe, clearly evidences the  benefits of fusing 2D joint location confidence maps with 3D 
image  cues, as done by our approach. Furthermore,  we also achieve lower error than
the  method  of~\cite{Rogez16},  despite  the fact  that  it  relies  on
additional training data. Even though our algorithm uses
only individual  images, it  also outperforms the  methods that  rely on
sequences~\cite{Du16,Tekin16a,Zhou16a}.  

\pf{
Since results are reported in~\cite{Bogo16,Sanzari16,Pavlakos16} for the average accuracy over all actions using 
	the Procrustes transformation, as explained in Section~\ref{ssec:eval}, 
	we do the same when comparing against these methods. Table~\ref{tab:procrustes} shows that we also outperform these baselines.
}

\comment{
As explained in Section~\ref{ssec:eval},~\cite{Bogo16} and~\cite{Sanzari16} report pose estimation results using 
Procrustes transformation. We carried out the same experiment and report results in Table~\ref{tab:procrustes}.
\bt{The parallel work of~\cite{Pavlakos16} adapts two different evaluation protocols.
The first protocol aligns the prediction to the ground-truth by Procrustes transformation. In the second one, 
as we confirmed through private communication, it is assumed that the depth of root joint position is known and 
pose of the person in the metric space is recovered from the volumetric representation using ground-truth z-axis 
of the root joint. By contrast, we do not assume that the ground-truth depth of the person is known. For a fair 
comparison, we also examine the performance of our approach against~\cite{Pavlakos16} using Procrustes transformation.}
}

\vspace{-0.4cm}
\paragraph{HumanEva.}

In Table~\ref{tab:humaneva}, we present the performance of our fusion approach on the HumanEva-I 
dataset~\cite{Sigal06}. We adopted the evaluation protocol described in~\cite{Bogo16,Simo-Serra13,Yasin16,Zhou16a} 
for a fair comparison. As in~\cite{Bogo16,Simo-Serra13,Yasin16,Zhou16a}, we measure 3D pose error as the average joint-to-joint 
distance after alignment by a rigid transformation.
Our approach also significantly outperforms the state-of-the-art on this dataset.

\begin{table}[t]
\centering
\tabcolsep=0.1cm
\scalebox{0.82}{
	\begin{tabular}[b]{lc}
		\toprule
		Method:												& 3D Pose Error \\
		\midrule
		Sanzari et al.~\cite{Sanzari16}		  				& 93.15		             \\
		Bogo et al.~\cite{Bogo16}			  				& 82.3		 		        \\
		Pavlakos et al.~\cite{Pavlakos16} 					& 53.2						\\
		\midrule
		Ours  						  						& \textbf{50.12}	 \\
		\bottomrule
	\end{tabular}
}
\caption{Comparison of our approach to the state-of-the-art methods that use
	Procrustes transformation on Human3.6m. We report 3D joint position errors (in mm).}
\label{tab:procrustes}
\end{table}

\begin{table}[tbph]
	\centering
	\tabcolsep=0.3cm
	\scalebox{0.82}{
		\begin{tabular}[b]{lcccc}
			\toprule
			Method 									& S1		 & S2         & S3     		& Average   \\
			\midrule
			Simo-Serra et al.~\cite{Simo-Serra13}  	& 65.1		 & 48.6		  &73.5	 	  	& 62.4			\\
			Bogo et al.~\cite{Bogo16}				& 73.3		 & 59.0		  &99.4			& 77.2				\\
			Zhou et al.~\cite{Zhou16a} 				& 34.2		 & 30.9		  &49.1			& 38.07				\\
			Yasin et al.~\cite{Yasin16}				& 35.8       & 32.4       &41.6         & 36.6              \\
			Tekin et al.~\cite{Tekin16a}			& 37.5       & 25.1       & 49.2        & 37.3           \\
			Ours					  			    & {\bf 27.24}& {\bf 14.26}&{\bf 31.74}  & {\bf 24.41}      		\\
			\bottomrule
		\end{tabular}
	}
	\caption{Quantitative results of our fusion approach on the Walking sequences of the HumanEva-I dataset~\cite{Sigal06}. S1, S2 and S3 correspond
		to Subject 1, 2, and 3, respectively. The accuracy is reported in terms of average Euclidean distance (in mm) between the
		predicted and ground-truth 3D joint positions.}
	\vspace{-4mm}
	\label{tab:humaneva}
\end{table}

\paragraph{KTH Multiview Football.}
In  Table~\ref{tab:kth}, we  compare our  approach  to~\cite{Belagiannis14a,Burenius13,Pavlakos16,Tekin16a} on the KTH  Multiview Football  II
dataset. Note that~\cite{Belagiannis14a} 
and~\cite{Burenius13}  rely on multiple views, and~\cite{Tekin16a} makes use of video data. 
As  discussed  in Section~\ref{ssec:eval}, we report the results of two instances of our 
model: one trained on the standard KTH training data, and one pretrained on the synthetic 3D 
human pose  dataset of~\cite{Chen16} and  fine-tuned on  the KTH  dataset. Note that, while 
working with a single input image, both instances outperform all the baselines. Note also that 
pretraining on synthetic data yields the highest accuracy. We believe that this further demonstrates    
the generalization ability of our method. 

In Fig.~\ref{fig:results}, we  provide representative poses predicted by our approach on
the Human3.6m, HumanEva and KTH Multiview Football datasets.

\begin{table}[t]
	\centering
	\tabcolsep=0.1cm
	\scalebox{0.62}{
	\begin{tabular}[b]{lccccccc}
		\toprule
		
		Method:         &\cite{Burenius13}    &\cite{Burenius13} &\cite{Belagiannis14a}  & \cite{Tekin16a} & \cite{Pavlakos16}& Ours-NoPretraining     & Ours-Pretraining   \\
		Input:			& Image				  &Image 		     & Image          		 & Video           & Image			  & Image    			   & Image    \\
		Num. of cameras:&1   				  &2 				 &2  					 &1  	    	   & 1				  & 1     			       & 1 \\
		\midrule
		Pelvis  	    &97    				  &97				 &-						 &99       		   & -			      &66					   & {\bf100}   \\
		Torso		    &87					  &90 				 &-						 &{\bf 100} 	   & -			      &{\bf100}				   & {\bf100}  \\
		Upper arms      &14 				  &53  				 &64     				 &74       		   & 94				  &74					   & {\bf100}  \\
		Lower arms      &06					  &28				 &50					 &49       		   & 80			      &{\bf100}			       & 88  		\\
		Upper legs      &63					  &88				 &75 					 &98  	  		   & 96				  &{\bf100}	    		   & {\bf100}  \\
		Lower legs      &41				   	  &82				 &66 					 &77	      	   & 84				  &77					   & {\bf88}  	\\
		All parts       &43					  &69				 &-						 &79  	  	       & -				  &83.2					   & {\bf95.2}   \\
		\bottomrule
	\end{tabular}}
	\caption{On KTH Multiview Football  II, we compare our method
        that uses a single image to those of~\cite{Burenius13,Pavlakos16,Tekin16a} 
        that use either one or two images, the one of~\cite{Belagiannis14a}  
        that  uses  two, and  the  one of~\cite{Tekin16a} that operates 
        on a sequence. As in~\cite{Belagiannis14a,Burenius13,Pavlakos16,Tekin16a}, we measure performance as the percentage of correctly estimated parts (PCP)
		score.
		A higher PCP score corresponds to better 3D pose estimation accuracy.}
	\label{tab:kth}
	\vspace{-2mm}
\end{table}

\subsection{Detailed Analysis}
\label{ssec:analysis}
\begin{table}[t]
	\centering
	\tabcolsep=0.1cm
	\scalebox{0.88}{
		\begin{tabular}[b]{lc}
			\toprule
			Method:  						 					& 3D Pose Error            \\
			\midrule
			Image-Only						  					& 124.13  	 	  			\\
			CM-Only 											& 79.28                          \\
			Early Fusion									    & 76.41   	  	 	            \\
			Late Fusion						  					& 74.12 		                    \\
			Trainable Fusion  						  			& {\bf 69.73}		                    \\
			\bottomrule
		\end{tabular}
	}\vspace{-1mm}
	\caption{Comparison of different fusion strategies and single-stream baselines on Human3.6m. We report the 3D joint position errors (in  mm).  The fusion
		networks perform better than those  that use only the image or
		only the confidence  map as input. Our trainable fusion achieves the
		best accuracy overall.}
	\vspace{-5mm}
	\label{tab:baseline_results}
\end{table}

\begin{figure}[t]
	\centering
	\scalebox{0.84}{
		\begin{tabular}{cc}
			\includegraphics[width=0.45\columnwidth]{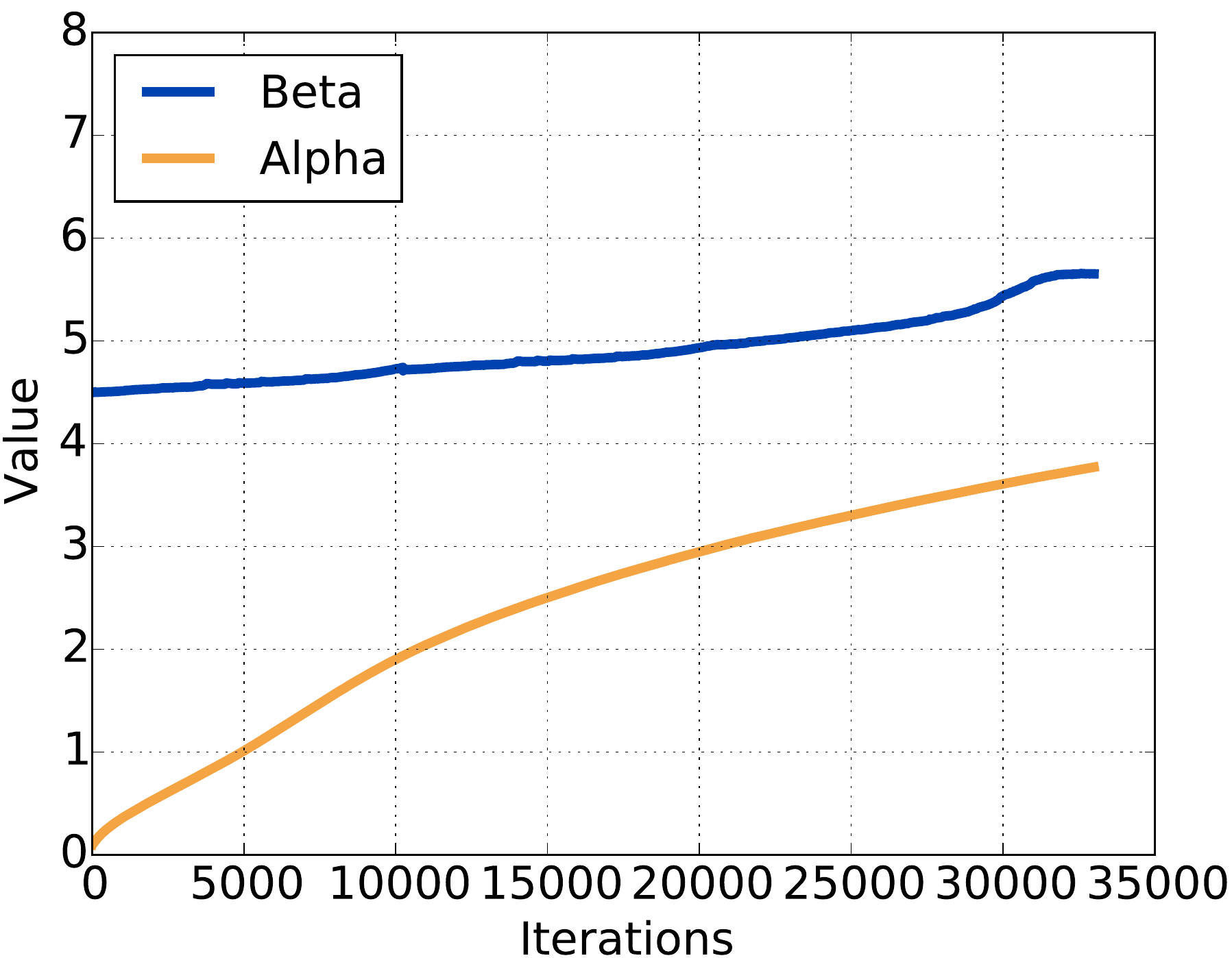}  
			& \includegraphics[width=0.45\columnwidth]{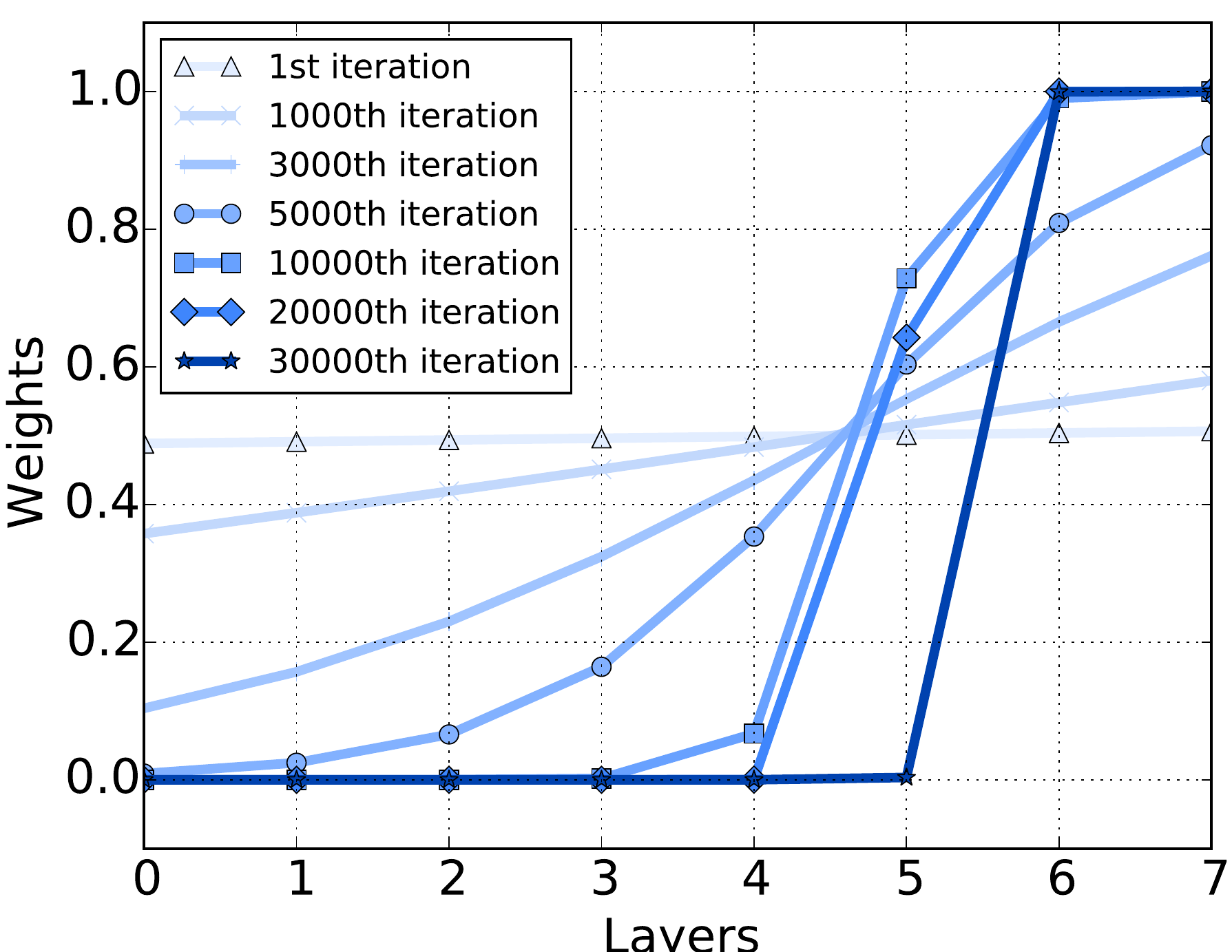}  \\
			\includegraphics[width=0.45\columnwidth]{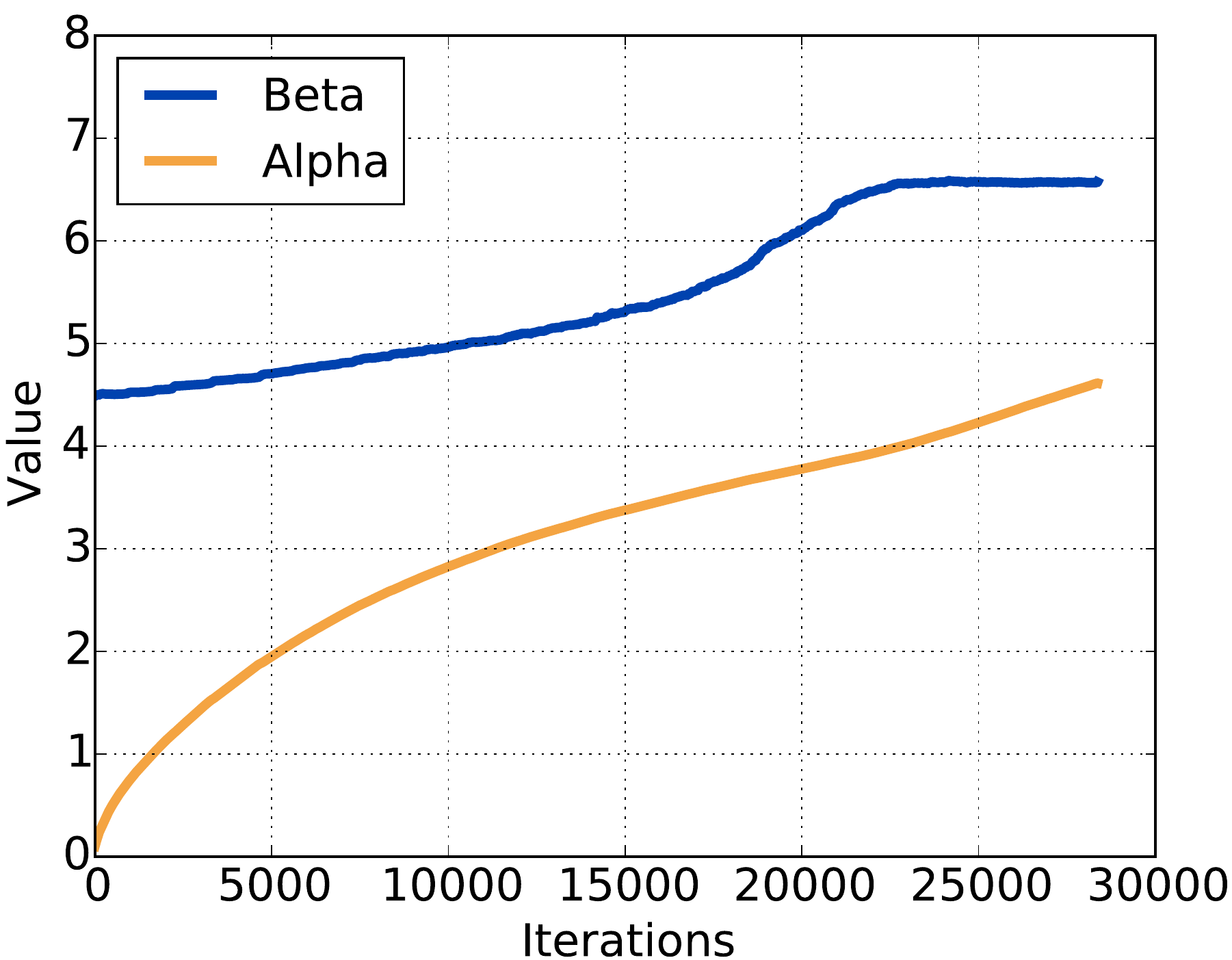}  
			& \includegraphics[width=0.45\columnwidth]{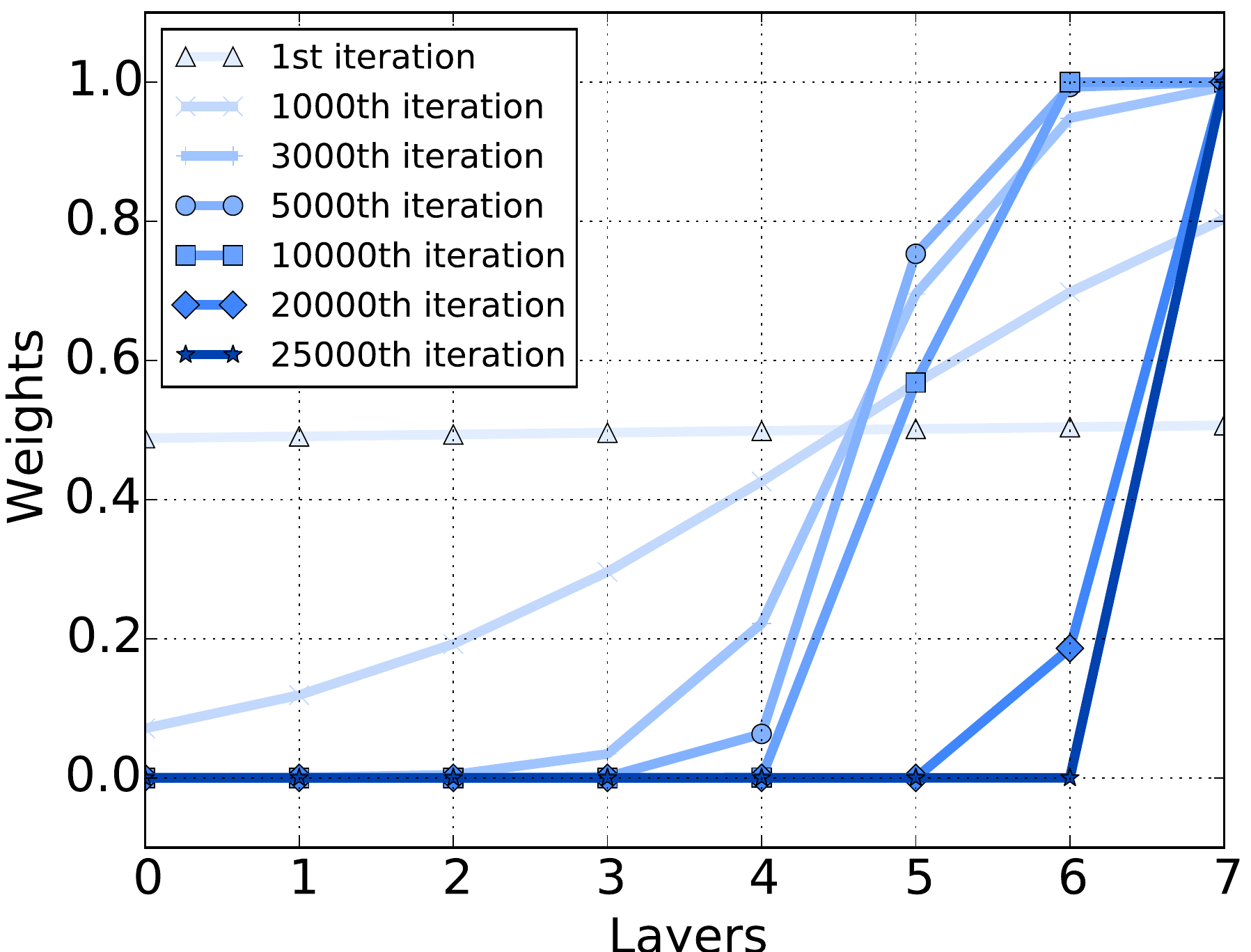}  \\
			\includegraphics[width=0.45\columnwidth]{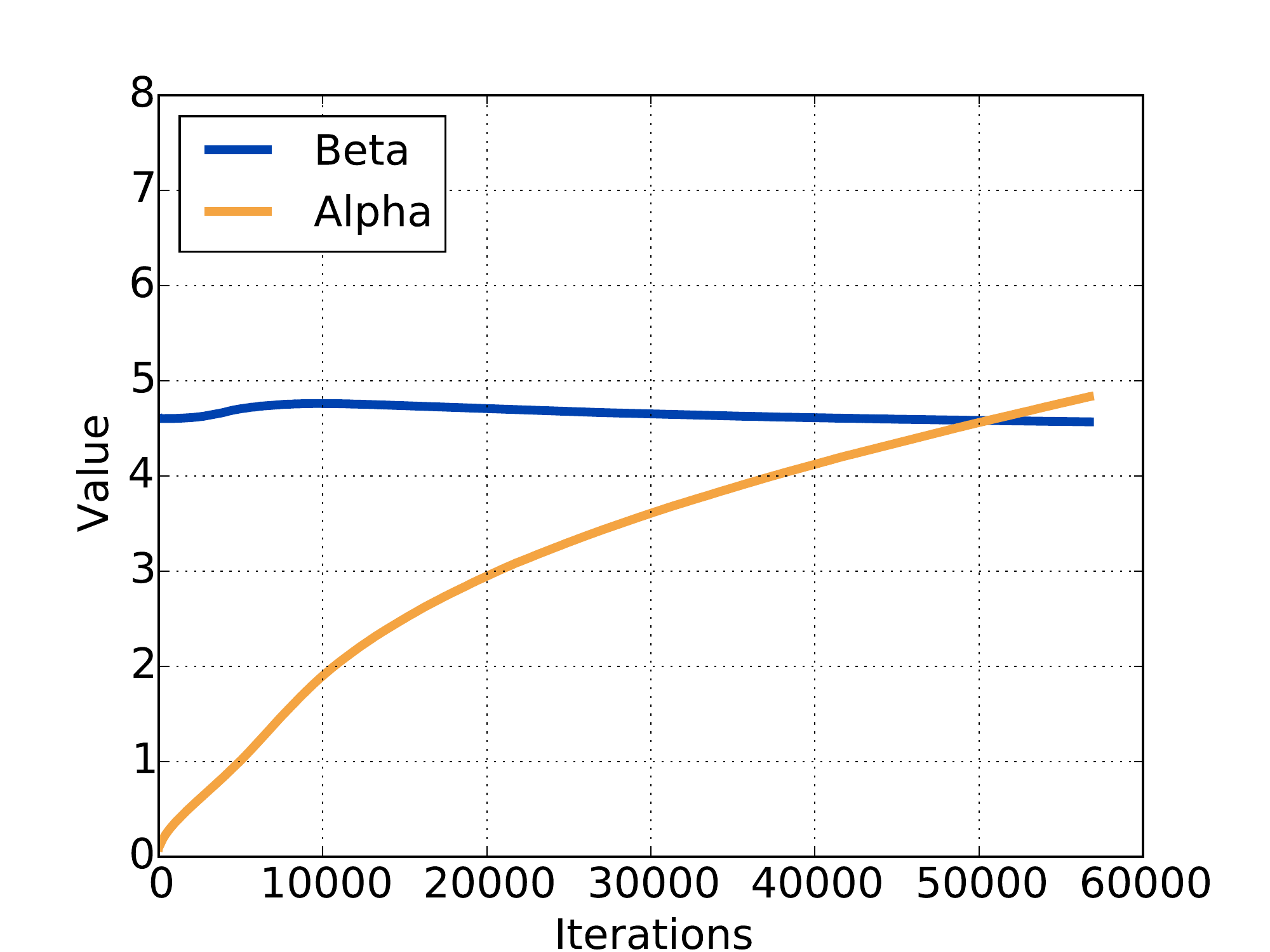}  
			& \includegraphics[width=0.45\columnwidth]{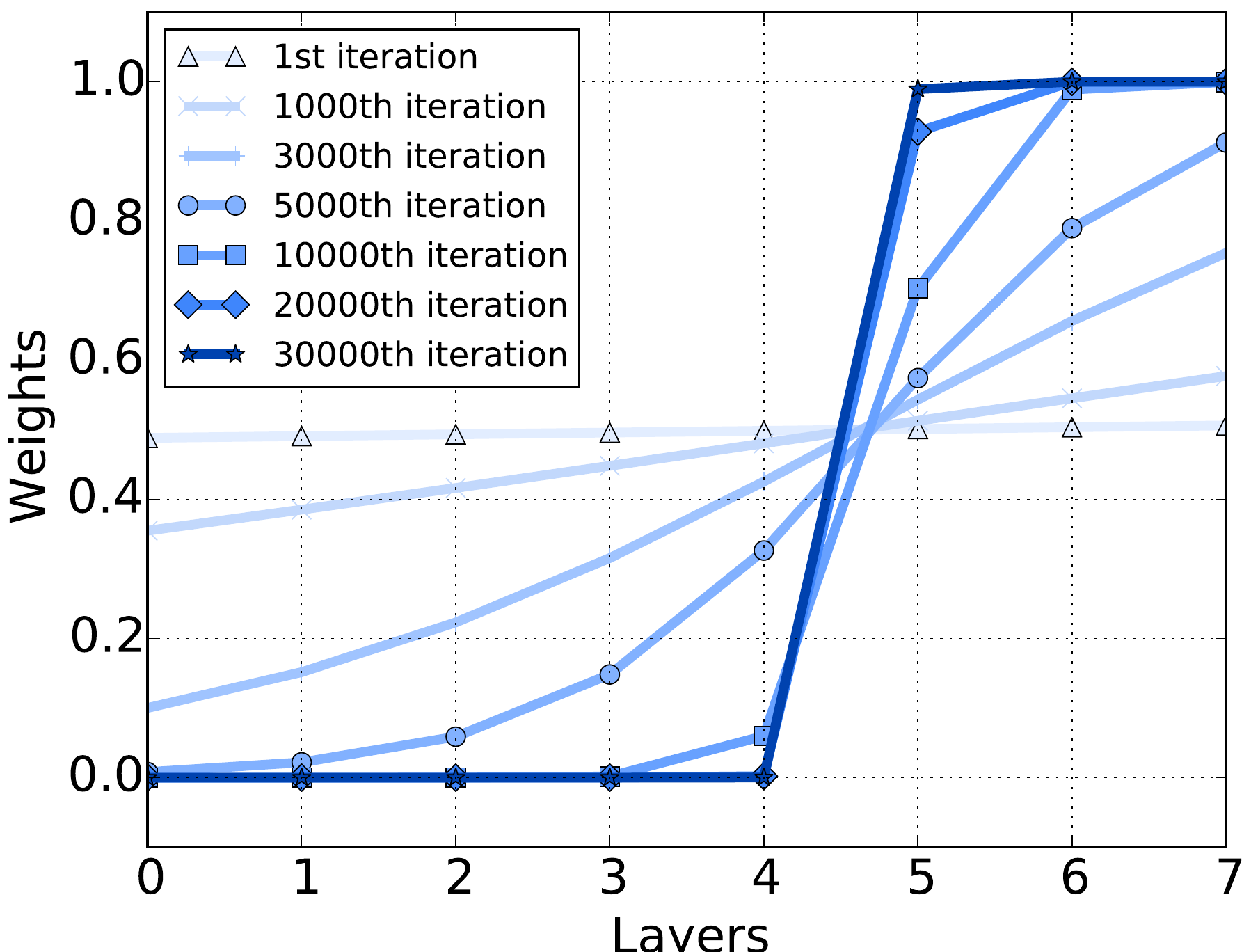}  \\
			\multicolumn{1}{c}{\footnotesize (a)}&\multicolumn{1}{c}{\footnotesize (b)}
		\end{tabular}
	}\vspace{-4mm}
	\caption{Evolution of {\bf(a)}~$\alpha$ and $\beta$, and {\bf(b)} the fusion weights in Human3.6m as training progresses. 
		Top row: Directions; Middle row: Discussion; Bottom row: Sitting Down.}
	\label{fig:weights}
	\vspace{-6mm}
\end{figure}

\begin{figure*}[t]
	\centering
	\scalebox{0.8}{
		\begin{tabular}{cc}
			\includegraphics[width=0.45\linewidth]{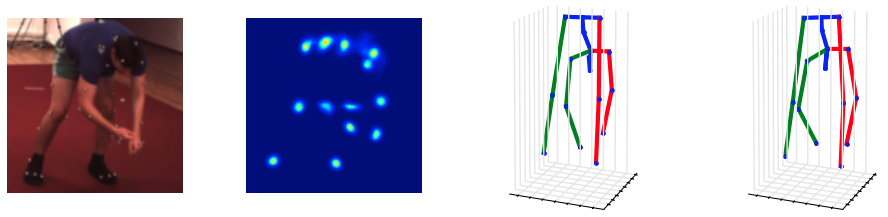} \hspace{6mm}
			&\includegraphics[width=0.45\linewidth,height=1.09in]{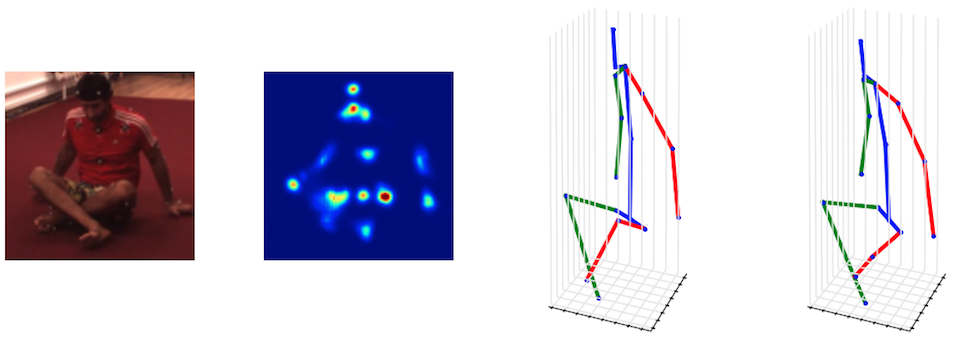} \\
			\includegraphics[width=0.45\linewidth,height=1in]{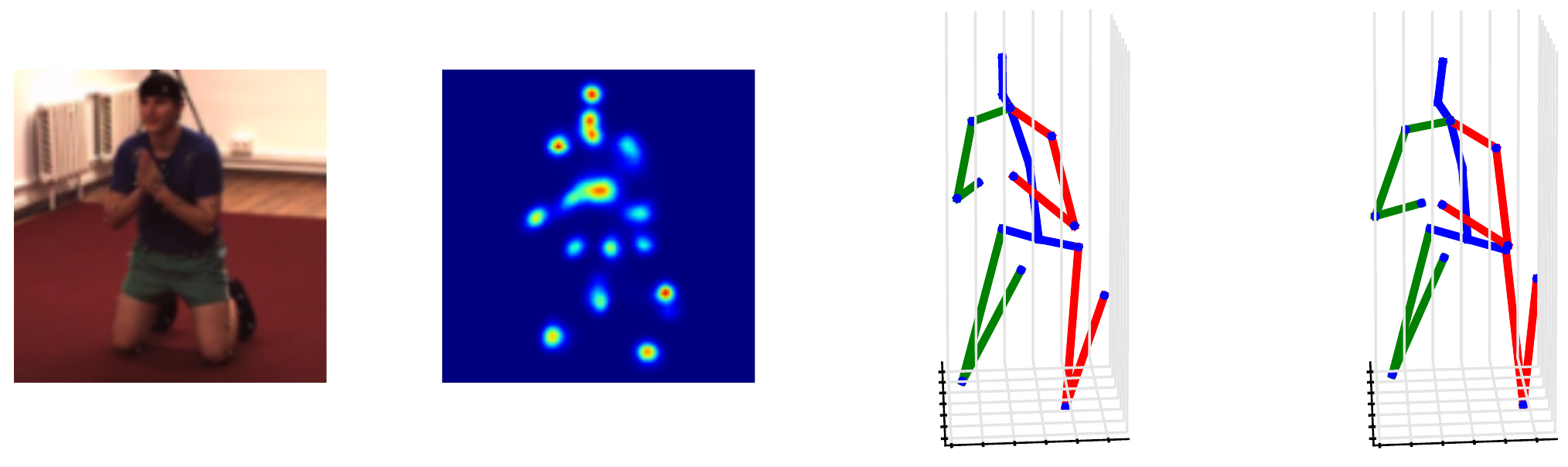} \hspace{6mm}
			& \includegraphics[width=0.45\linewidth,height=1in]{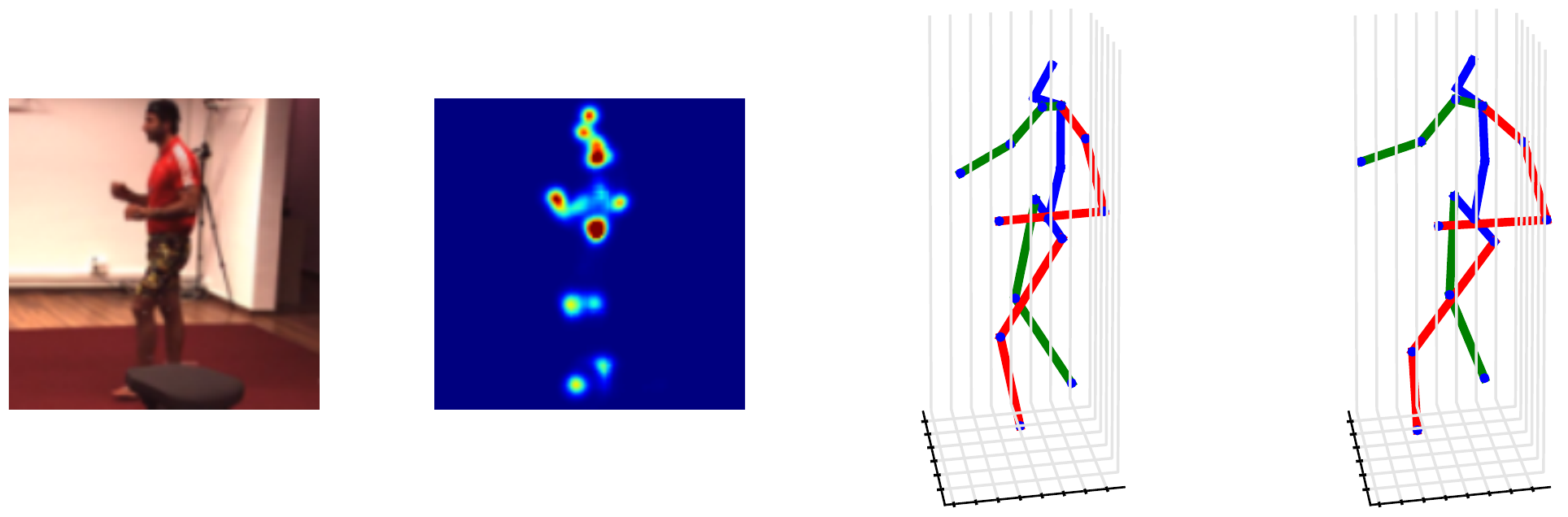} \\
			\includegraphics[width=0.45\linewidth]{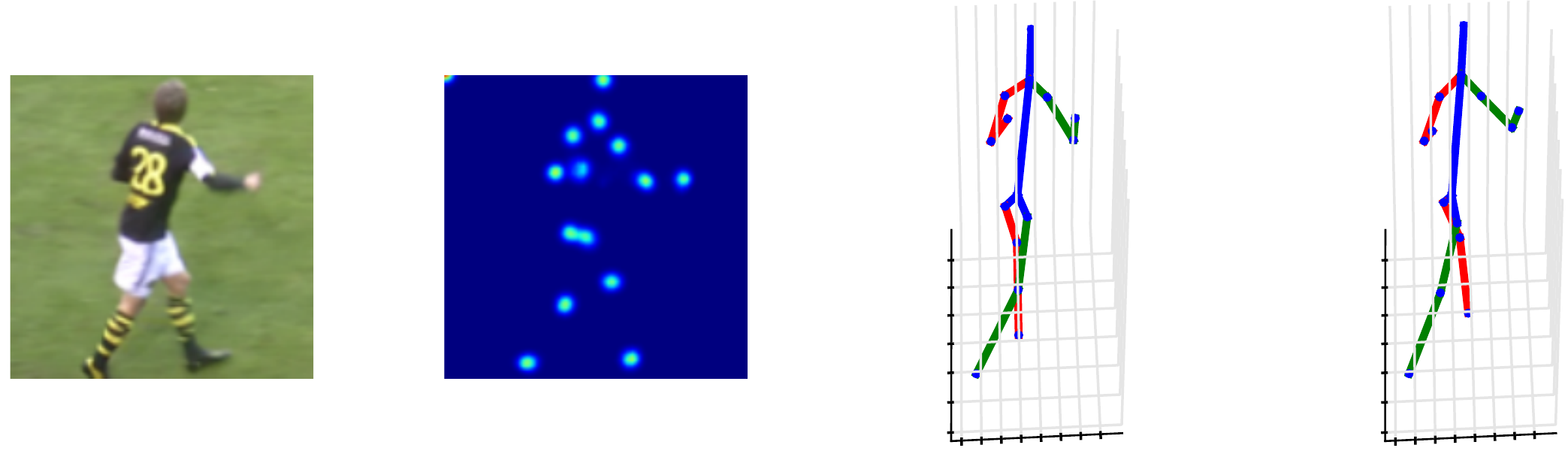} \hspace{6mm}
			& \includegraphics[width=0.45\linewidth]{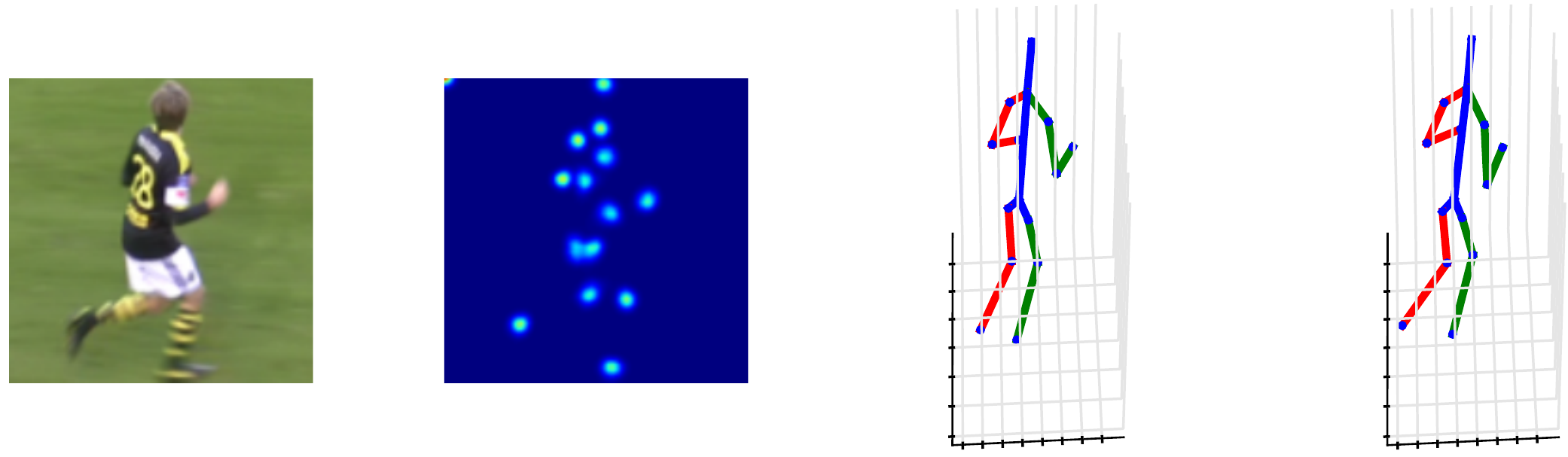} \\
			\includegraphics[width=0.45\linewidth]{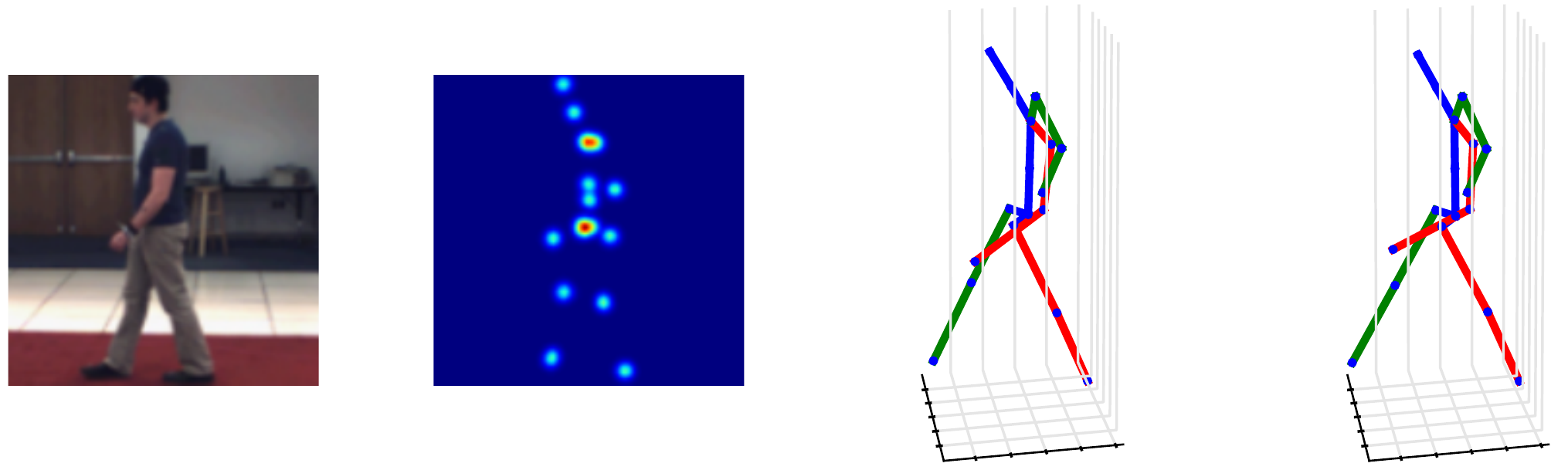} \hspace{6mm}
			& \includegraphics[width=0.45\linewidth]{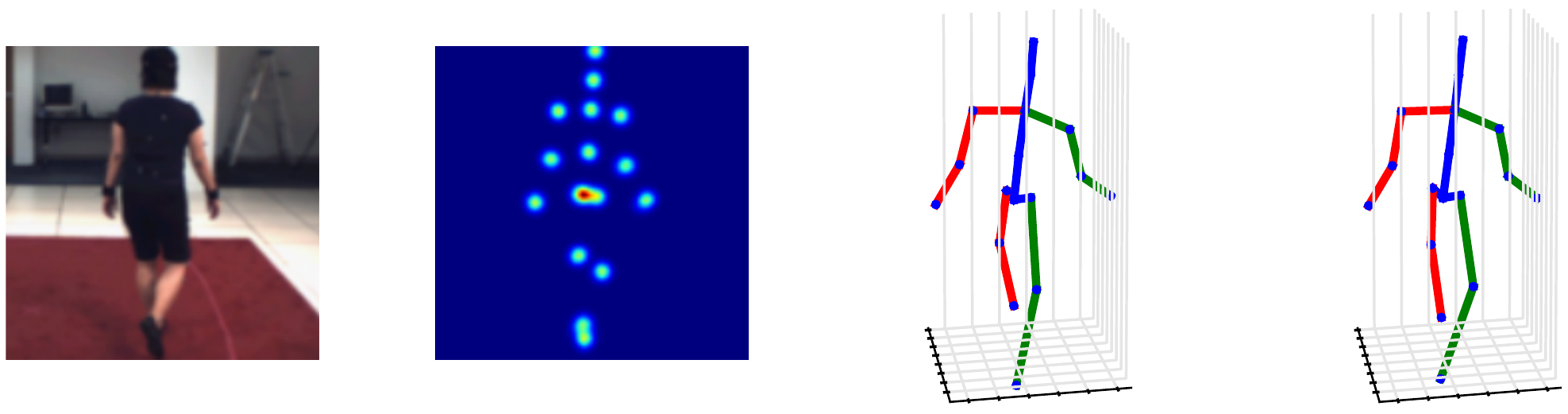} \\
			\footnotesize (a) Image \;\;\; (b) Confidence Map \;\;\; (c) Prediction \;\;\; (d) Ground-truth & \footnotesize \;\;\;\;\;\;\;\;\;\;\; (e) Image \;\;\; (f) Confidence Map \;\;\; (g) Prediction \;\;\; (h) Ground-truth \;\;\;
		\end{tabular}
	} \vspace{0mm}
	\caption{{\bf Pose  estimation  results  on  Human3.6m, HumanEva and KTH Multiview Football.} {\bf  (a, e)}  Input
		images. {\bf (b, f)} 2D joint  location confidence maps.
		{\bf (c, g)} Recovered pose. {\bf (d, h)} Ground truth.
		Note that our
		method can  recover  the  3D  pose  in these  challenging  scenarios,  which
		involve significant amounts of self occlusion  and orientation ambiguity. Best viewed in color.}
	\label{fig:results}
\end{figure*}

We now analyze two different aspects of our approach. First, we compare our trainable
fusion approach to early fusion, depicted in Fig.~\ref{fig:hardcoded}(a), and late fusion, depicted in
Fig.~\ref{fig:hardcoded}(c). Then, we analyze the benefits 
of leveraging both 2D joint locations with their corresponding uncertainty and additional image cues. 
To this end, we make use of two additional baselines. The first one consists of 
a single stream CNN regressor operating on the image only. We refer to this baseline as \emph{Image-Only}. 
The second is a CNN trained to predict 3D pose from only the 2D 
confidence map (CM) stream. We refer to this baseline as \emph{CM-Only}.

In Table~\ref{tab:baseline_results}, we report the average pose estimation errors on Human3.6m
for all these methods. Our trainable fusion strategy yields the best results. Note 
also that, in general, all fusion strategies outperform the state-of-the-art methods in Table~\ref{tab:overall}. 
Importantly, the \emph{Image-Only} and \emph{CM-Only} baselines perform worse than our approach, 
and all fusion-based methods. This demonstrates the importance of fusing 2D joint location confidence
maps along with 3D cues in the image for monocular pose estimation.

\ms{In Fig.~\ref{fig:weights}, we depict the evolution throughout the training iterations of {\bf (a)}~the
parameters $\alpha$ and $\beta$ that define the weight vector in our trainable fusion framework
as given by Eq.~\ref{eq:sig}, 
and {\bf (b)} the weight vector itself.
%
%
%
An increasing value of $\alpha$, expected due 
to our regularizer, indicates that fusion becomes sharper throughout the training. An increasing $\beta$, which is 
the typical behavior, corresponds to fusion occurring in the later stages of the network.
We conjecture 
that this is due to the fact that features learned by the image and confidence map streams at later
layers become less correlated, and thus yield more discriminative power. \comment{To analyze this, we have measured 
the correlation between the features of the confidence map stream and those of the image stream after the 
last convolutional layer of the network for our trainable fusion strategy and compared it against the 
correlation of the feature maps of the early fusion scheme in terms of Pearson correlation coefficient values. 
\comment{\MS{What is this?}}  As can be seen in Fig.~\ref{fig:pearson}, where we visualize these values, the 
early fusion features are indeed more correlated than those obtained by our trainable fusion strategy, which 
typically tends to select a later layer to fuse the two streams.}}

\bt{
To analyze this further, we show in Fig.~\ref{fig:pearson}
the squared Pearson correlation coefficients between all pairs of features of the confidence map stream and
of the image stream at the 
last convolutional layer of our trainable fusion network. As can be seen in the figure, 
the image and confidence map streams produce decorrelated features that
are complementary to each other allowing to effectively account for different input
modalities.
}

\comment{
\MS{To be honest, I cannot see much from the figure...}}

\comment{
The sharp transitions of the weight vector decides where the fusion takes place. 
We have observed that fusion generally occurs later than earlier at the network. 
This might be due to the fact that high level features learned from the image and 
confidence map streams are more correlated at the early fusion, thus leading to more redundancy and less 
discriminative power. In order to analyze this further, we have measured 
the correlation between the feature maps of the confidence map stream and those of the image stream 
after the last convolutional layer of the network for our trainable fusion strategy
and compared it against the correlation of the feature maps of the early 
fusion in terms of $R^2$ values in Fig.~\ref{fig:r2}. While the feature maps of the early fusion are 
more correlated, the image and confidence map streams of trainable fusion scheme 
learns less correlated and complementary features that, when combined, improve
upon the general early fusion.
}

\subsection{Qualitative Results}

In Fig.~\ref{fig:leeds_examples}, we present qualitative pose estimation results on the Leeds Sports Pose
dataset. We trained our network on the synthetic dataset of~\cite{Chen16} and tested on images
acquired outdoors in unconstrained settings. The accurate 3D predictions of the challenging poses 
demonstrate the generalization ability and robustness of our method. 

\begin{figure}[t]
	\centering
	\scalebox{0.78}{
		\begin{tabular}{c}
			\includegraphics[width=\linewidth]{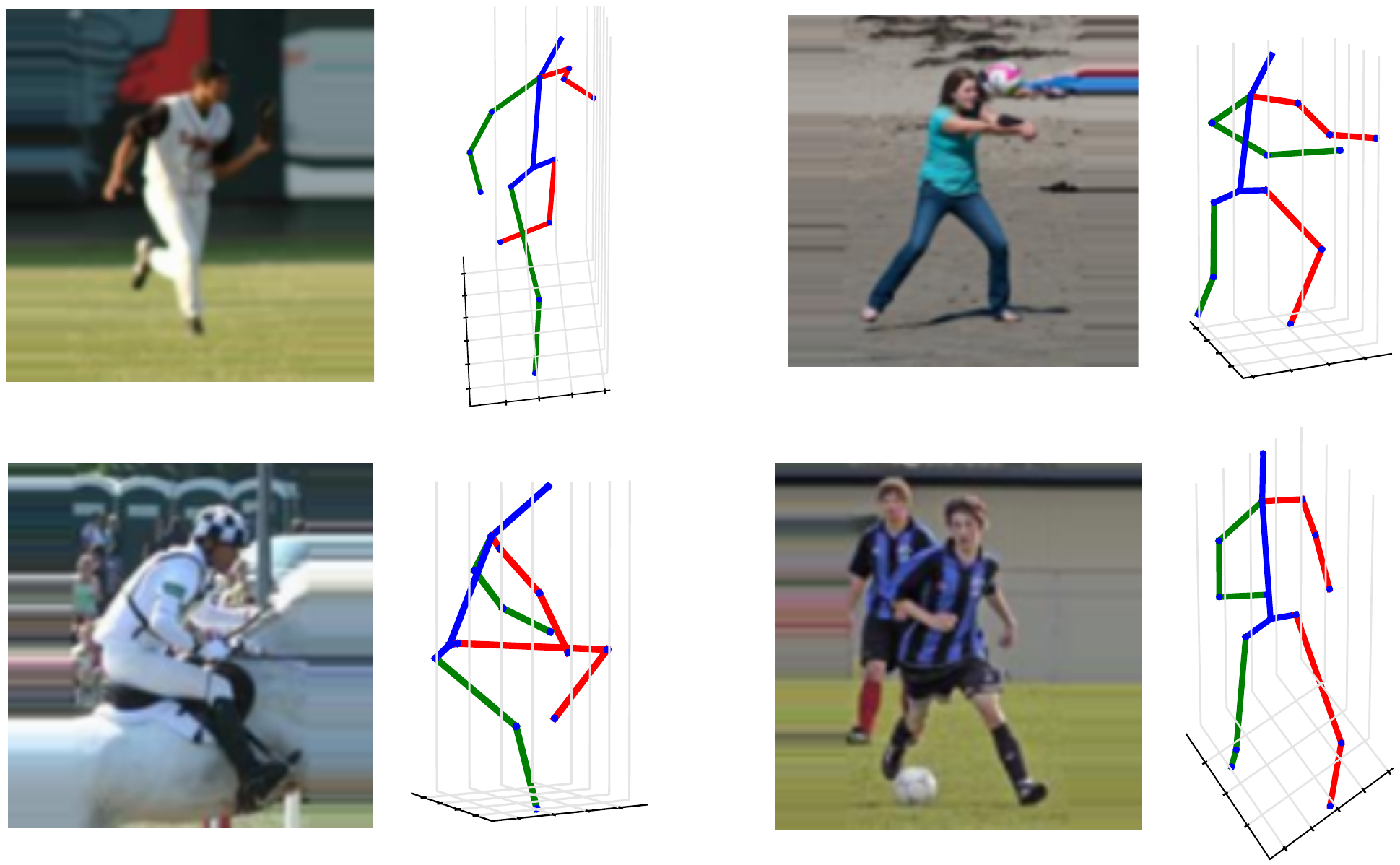}  \\
		\end{tabular}
	}
	\caption{{\bf Pose  estimation results  on the Leeds Sports Pose dataset.} We show
		the input image and the predicted 3D pose for four images. Best viewed in color.}
	\label{fig:leeds_examples}
	\vspace{-3mm}
\end{figure}

\begin{figure}[t]
	\centering
	\scalebox{0.77}{
	\begin{tabular}{c}
		\includegraphics[width=\columnwidth]{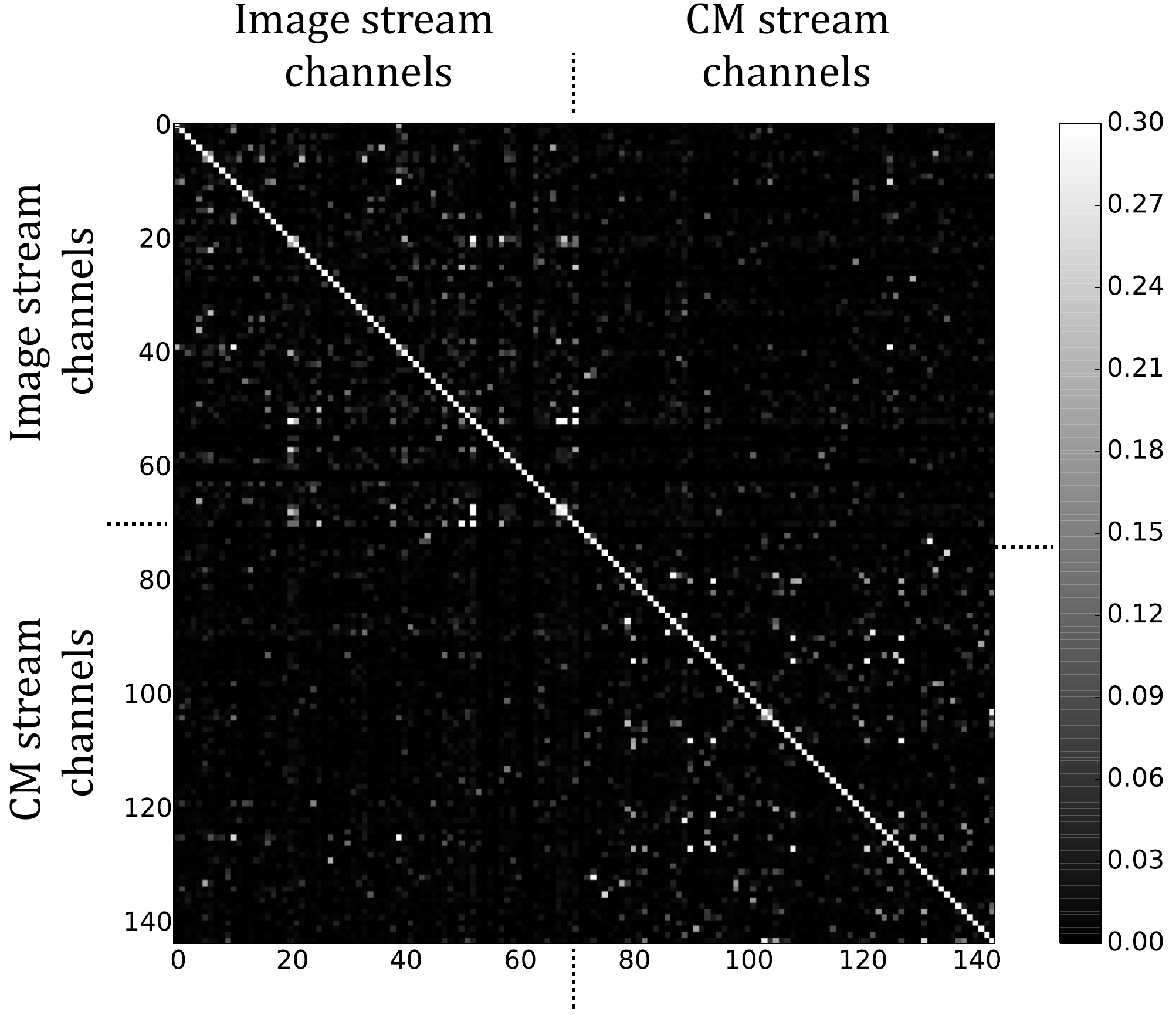}  
	\end{tabular}}
	\caption{Squared Pearson correlation coefficients ($R^2$) between each pair of the features
		learned at the last convolutional layer of our trainable fusion network computed
		from $128$~randomly selected images in Human3.6m. As can be seen in the lower left and
		upper right submatrices, the feature maps of the image and the 
		confidence map streams are decorrelated.}
	\label{fig:pearson}
	\vspace{-6mm}
\end{figure}

\section{Conclusion}

In this paper, we have proposed to fuse 2D and 3D image cues for monocular  3D  human pose  estimation.  To  this  end,  we have  introduced  an approach that relies on two CNN streams to jointly infer 3D pose from 2D joint locations and from the image directly. We have also introduced an approach to fusing the two streams in a trainable way. 

We have demonstrated that  the resulting CNN pipeline significantly outperforms  state-of-the-art methods on  standard 3D  human pose estimation  benchmarks.  Our  framework is general and can easily be extended to  incorporate other modalities, such as optical flow or body part segmentation. Furthermore, our trainable fusion strategy could be applied to other fusion problems, which is what we intend to do in future work. 

\comment{We have introduced a novel Deep Learning regression architecture for 3D human pose
estimation from a monocular image. We have shown that our deep fusion network
can account for both contextual 3D cues present in the image and the uncertainty
in joint locations. As a result, it yields better accuracy than the state-of-the-art on
standard 3D human pose estimation benchmarks. Our proposed framework is generic
and in future work, we intend to apply it to fuse contextual image information
with part segmentations and optical-flow in the context of human pose estimation
and deformable surface reconstruction.}

\comment{
compute representations of 2D joint location confidence
  maps and RGB images and automatically learns where to fuse these streams.} 

\comment{a two-stream Deep Network that computes  representations of 2D joint probability
  maps and RGB images,  and fuses them to predict 3D  pose. }
  
\comment{ In the future,
  we therefore intend  to study the influence of part  segmentations and optical
  flow  on human  pose  estimation, along  with  temporal consistency  when
  working with image sequences. \MS{Do we want to mention using the trainable fusion approach for different problems, such as action recognition, where many methods fuse different kinds of information, but currently manually determine where fusion occurs.}}

\appendix

\begin{figure*}[tbph]
	\centering
	\scalebox{0.9}{
		\begin{tabular}{cccccccc}
			\includegraphics[width=0.11\linewidth, height=3.2cm]{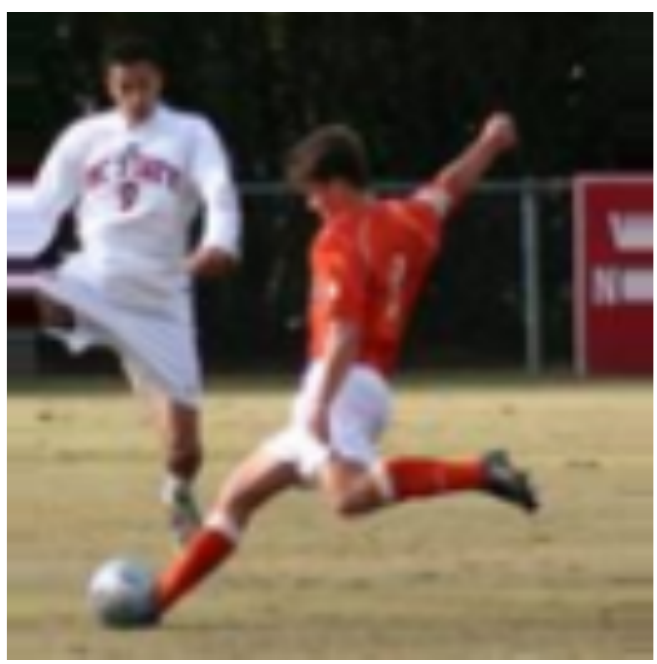} 
			&\includegraphics[width=0.11\linewidth, height=3.2cm]{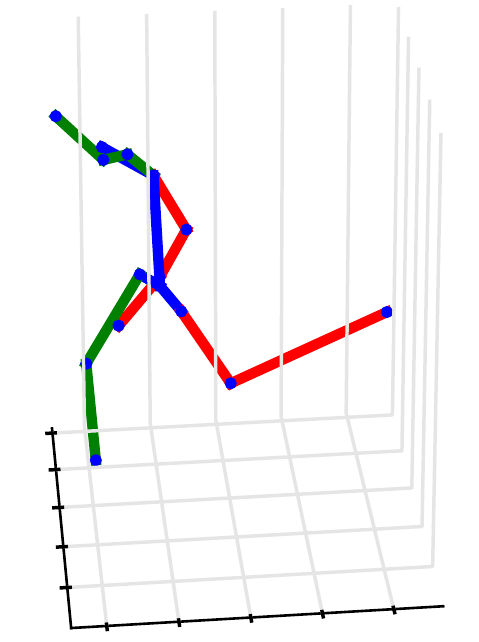} 
			&\includegraphics[width=0.11\linewidth, height=3.2cm]{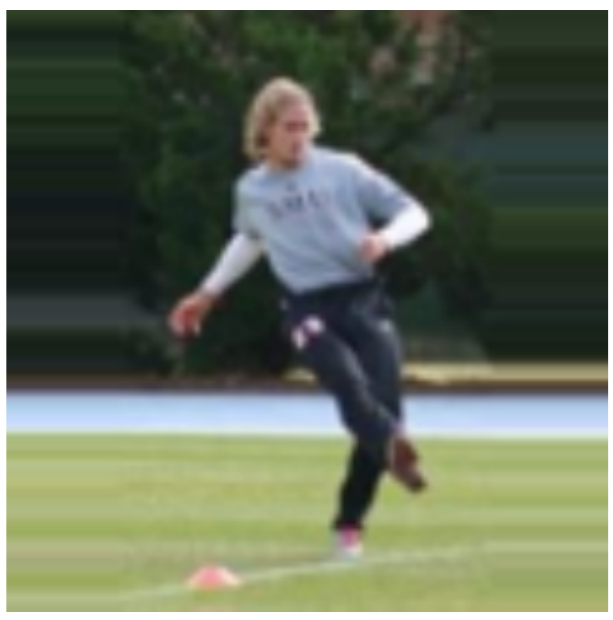} 
			&\includegraphics[width=0.11\linewidth, height=3.2cm]{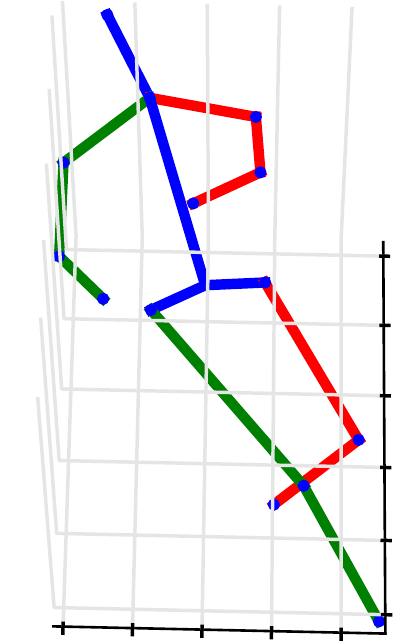}  
			&\includegraphics[width=0.11\linewidth, height=3.2cm]{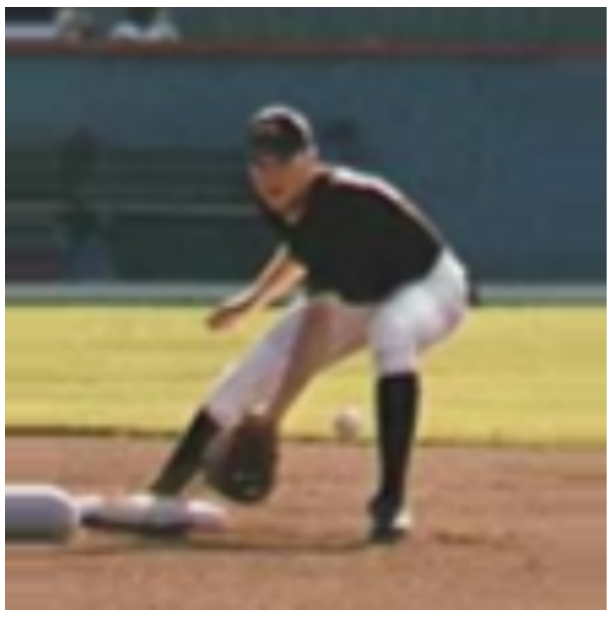} 
			&\includegraphics[width=0.11\linewidth, height=3.2cm]{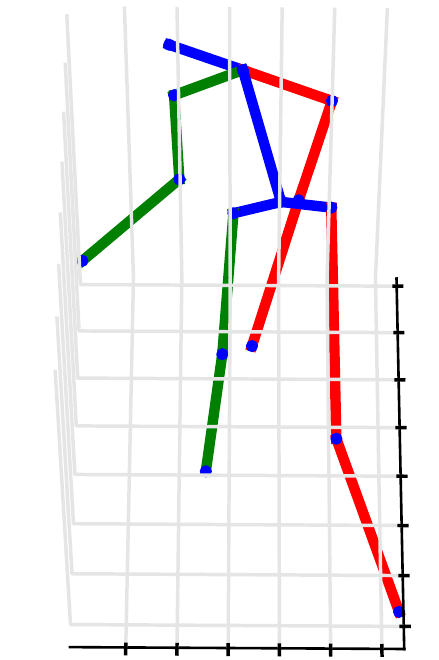} 
			&\includegraphics[width=0.11\linewidth, height=3.2cm]{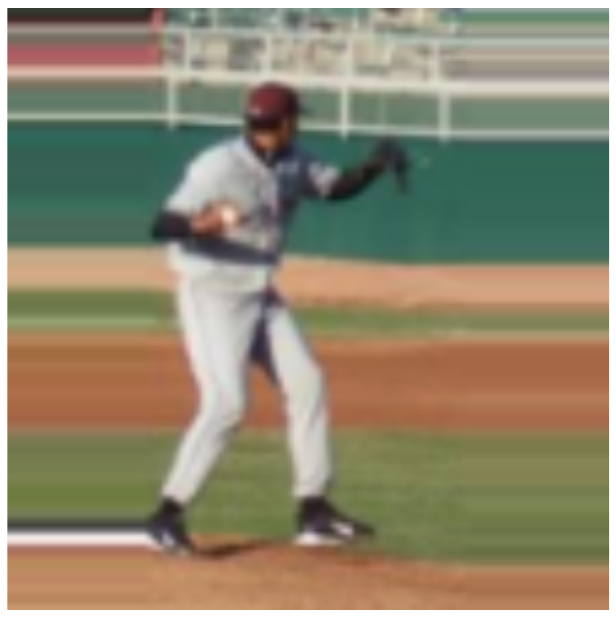} 
			&\includegraphics[width=0.11\linewidth, height=3.2cm]{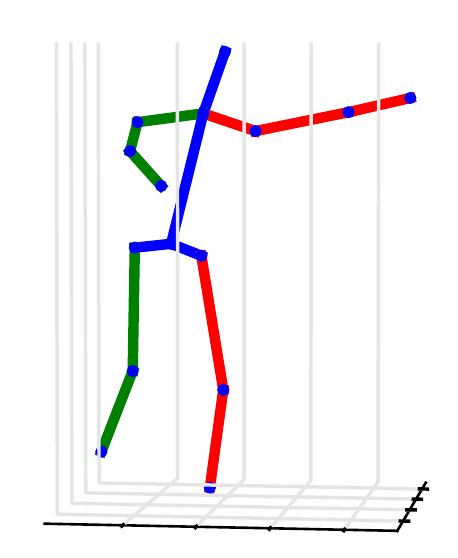}  \\ \vspace{6mm}
			\includegraphics[width=0.11\linewidth, height=3.2cm]{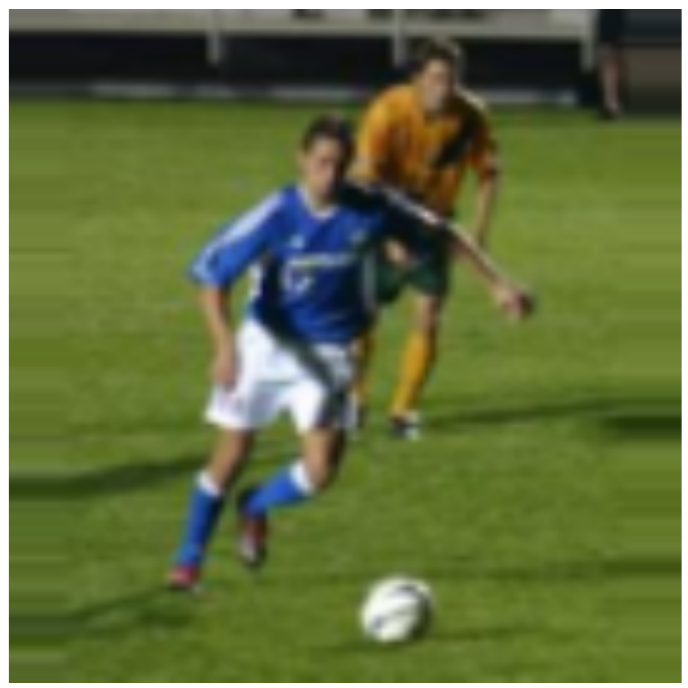} 
			&\includegraphics[width=0.11\linewidth, height=3.2cm]{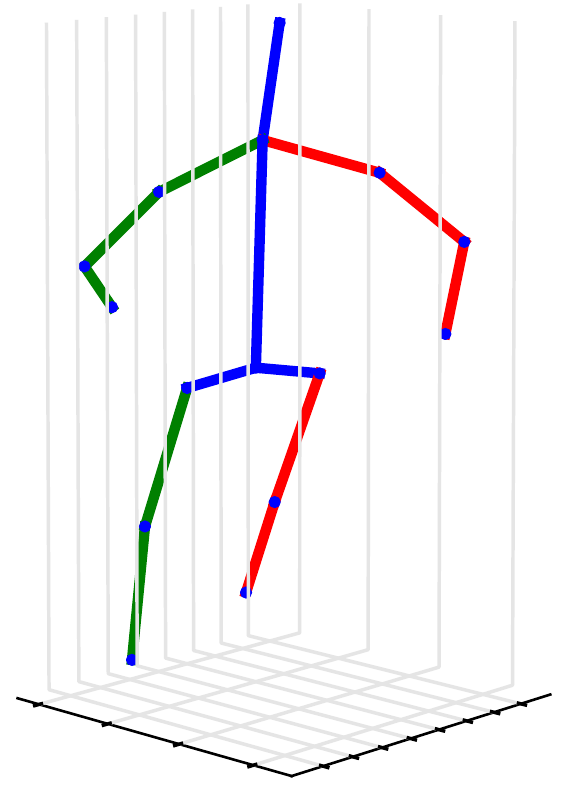} 
			&\includegraphics[width=0.11\linewidth, height=3.2cm]{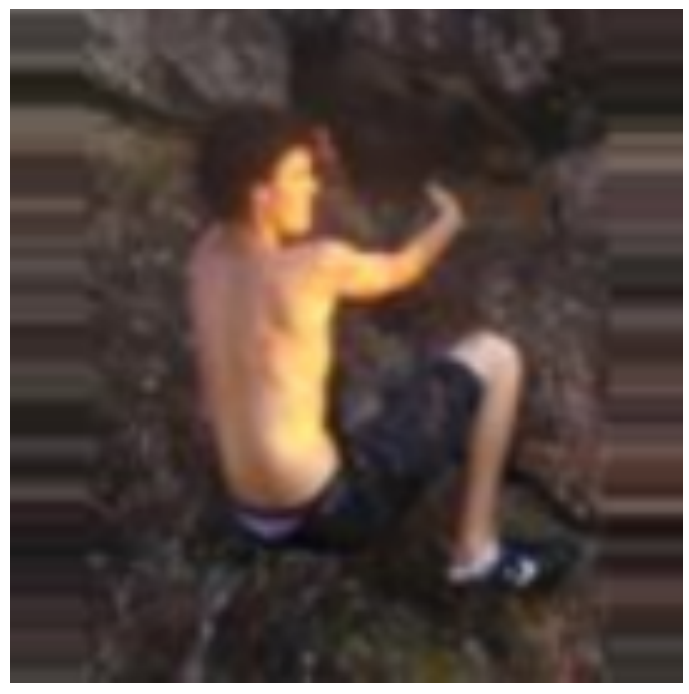} 
			&\includegraphics[width=0.11\linewidth, height=3.2cm]{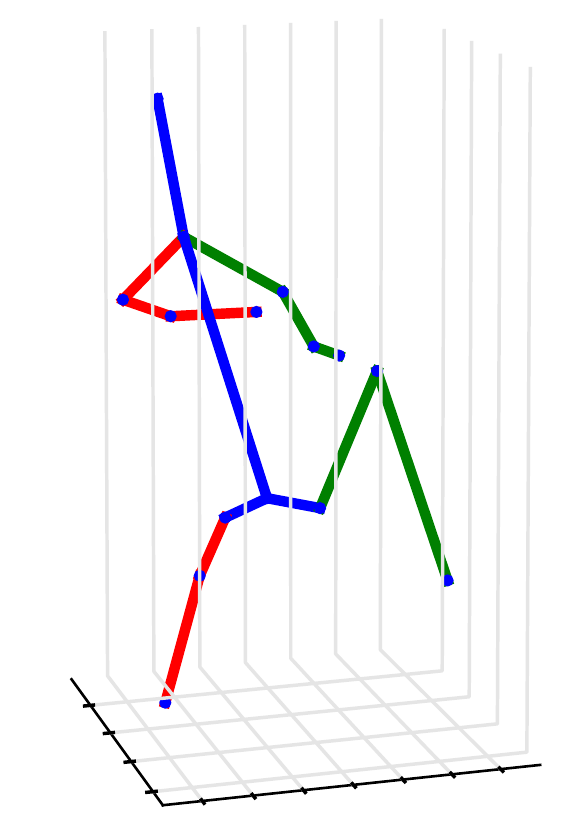}  
			&\includegraphics[width=0.11\linewidth, height=3.2cm]{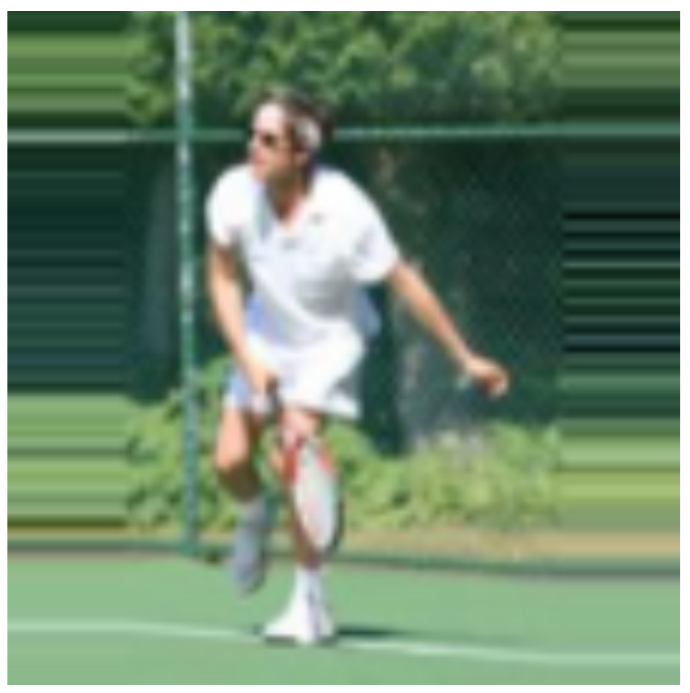} 
			&\includegraphics[width=0.11\linewidth, height=3.2cm]{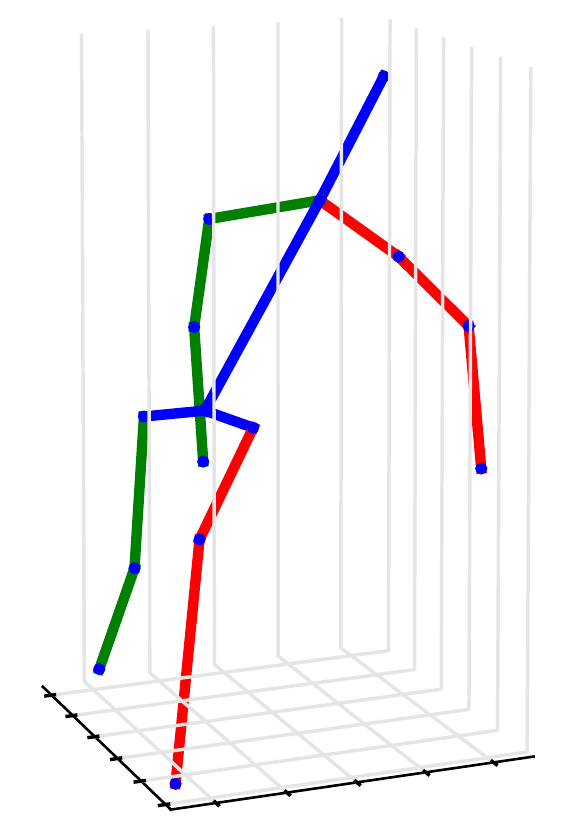} 
			&\includegraphics[width=0.11\linewidth, height=3.2cm]{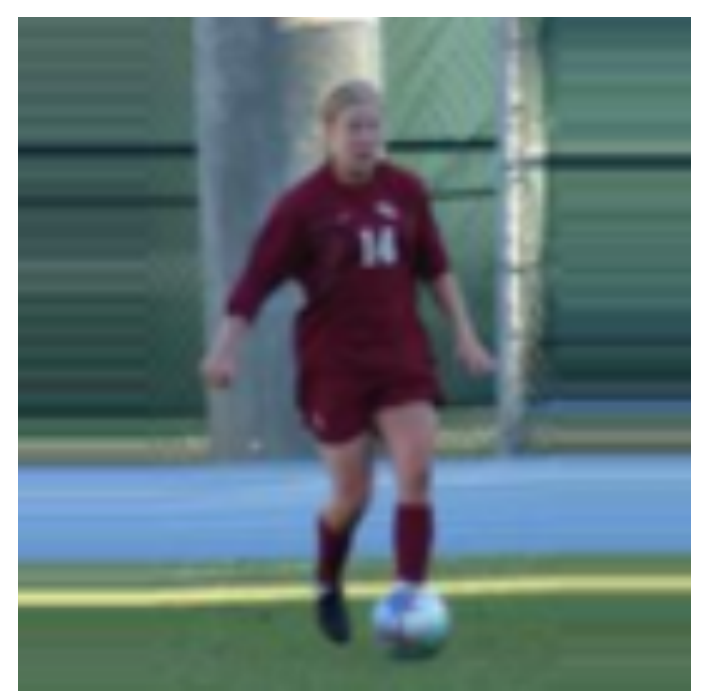} 
			&\includegraphics[width=0.11\linewidth, height=3.2cm]{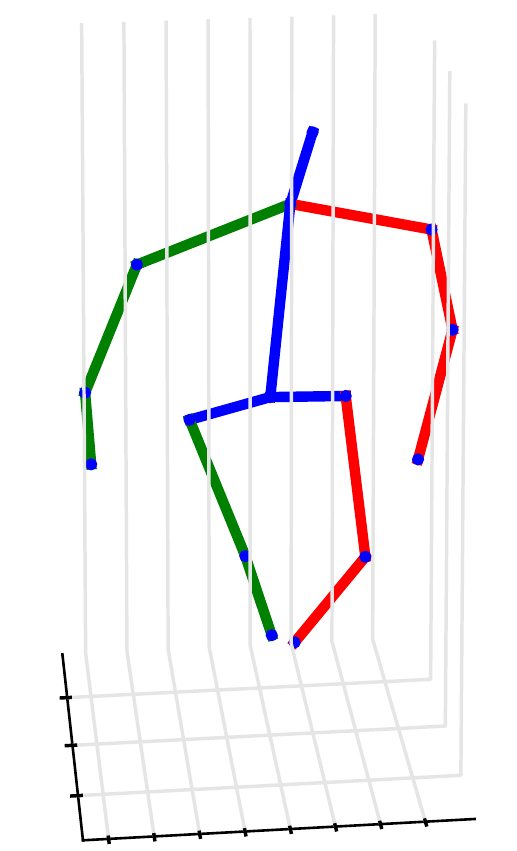}  \\
		\end{tabular}
	} 
	\caption{Pose  estimation  results  on  LSP. We trained our network 
		on the recently released synthetic dataset of~\cite{Chen16} and tested it
		on the LSP dataset. The quality of the 3D pose predictions 
		demonstrates the generalization of our method. Best viewed in color.}
	\vspace{3mm}
	\label{fig:results_lsp}
\end{figure*}

\section{Appendix}

In this appendix, we analyze the influence of our regularization 
term encouraging sharp fusion in Eq.~4, provide running time for our algorithm, and show additional qualitative results
on the Leeds Sports Pose~\cite{Johnson10b}, HumanEva-I~\cite{Sigal06}, Human3.6m~\cite{Ionescu14a} and KTH Multiview Football II~\cite{Burenius13}
datasets.

\vspace{2mm}

\paragraph{Effect of the regularization.}
Below, we analyze the effect of the regularization term that encourages sharp fusion in Eq.~4. In the absence
of the regularization term,
the network mixes the data and fusion streams without necessarily fusing them 
at a specific layer. As discussed in the main paper, this corresponds to a 
model with many active parameters. Therefore it is prone to overfitting and computationally 
less efficient at test-time. In Table~\ref{tab:reg_term}, we compare the results of our approach
with and without this regularization term. For the latter, we do not parametrize the weights of the network 
with a sigmoid function and do not constrain the network to have a sharp fusion.
The results confirm that encouraging sharp fusion yields both better accuracy 
and faster prediction.

\vspace{3mm}

\begin{table}[tbph]
	\centering
	\tabcolsep=0.3cm
	\scalebox{1}{
		\begin{tabular}[b]{lcccc}
			\toprule
			Method 									& 3D Pose Error		& Runtime           	   \\
			\midrule
			Without regularization				& 68.30		 		& 0.013					\\
			With  regularization  			    & 60.17				& 0.006    		       \\
			\bottomrule
		\end{tabular}
	}
	\caption{Quantitative results of our fusion approach with and without the regularization term encouraging sharp fusion. These experiments were carried out on 
		the \emph{Eating} action class of Human3.6m. 3D pose error is computed as the average Euclidean distance (in milimeters) between the predicted 
		and ground-truth 3D joint positions. Runtime denotes the computational time spent, in sec/frame, during testing for the fusion 
		network with and without the regularization term. With the regularization term, inactive layers are pruned after training, which yields a 
		more efficient network for test-time prediction.}
	\label{tab:reg_term}
\end{table}

\vspace{2mm}

\paragraph{Running time.}
We carried out our experiments on a machine equipped with an Intel Xeon CPU E5-2680 and an NVIDIA TITAN X Pascal GPU. 
It takes $90$ ms to compute 2D joint location confidence maps and $6$ ms to predict 3D pose with our fusion network.
Therefore, the total runtime of our method is 0.096 sec/frame (over 10 fps), which compares favorably with the 
recent model-based methods ranging from 0.04 fps to 1 fps~\cite{Yasin16,Sanzari16,Zhou16a}.

\vspace{2mm}

\paragraph{Additional qualitative results.}
We provide additional qualitative results for the HumanEva~\cite{Sigal06}, Human3.6m~\cite{Ionescu14a}, and KTH Multiview Football II~\cite{Burenius13} datasets in Figs.~\ref{fig:results_he}~\ref{fig:results_h36m}, 
and~\ref{fig:results_kth}, respectively. We further demonstrate that our regressor trained on the recently released synthetic dataset of~\cite{Chen16}
generalizes well to real images obtained from the Leeds Sports Pose dataset~\cite{Johnson10b} in Fig.~\ref{fig:results_lsp}. Additional 
qualitative results can be found in the accompanying videos.


\begin{figure*}[tbph]
	\centering
	\scalebox{1}{
		\begin{tabular}{cc}
			\includegraphics[width=0.45\linewidth, height=3.2cm]{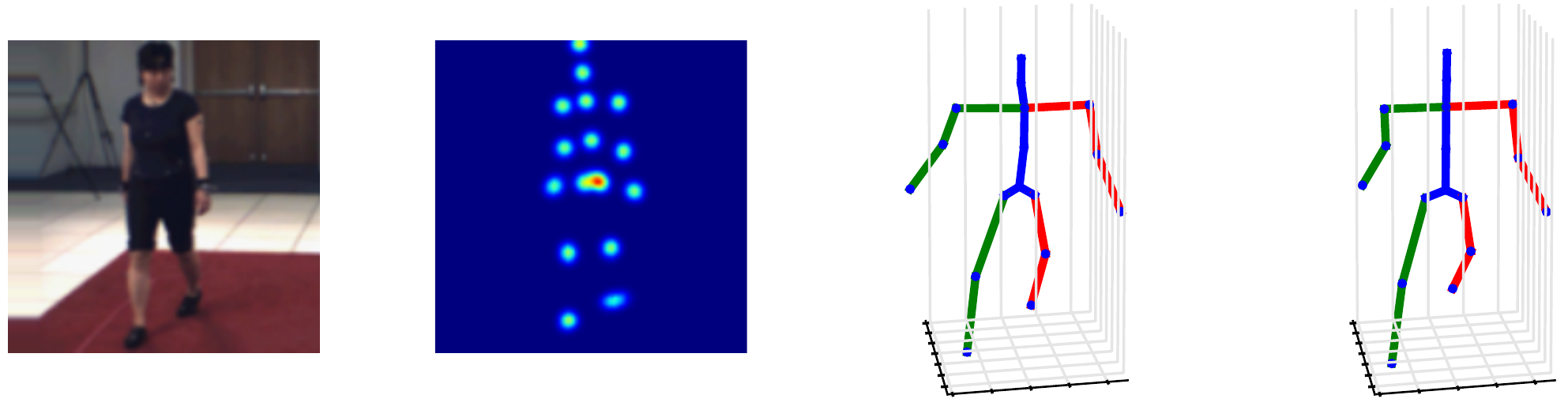} \hspace{6mm}
			& \includegraphics[width=0.45\linewidth, height=3.2cm]{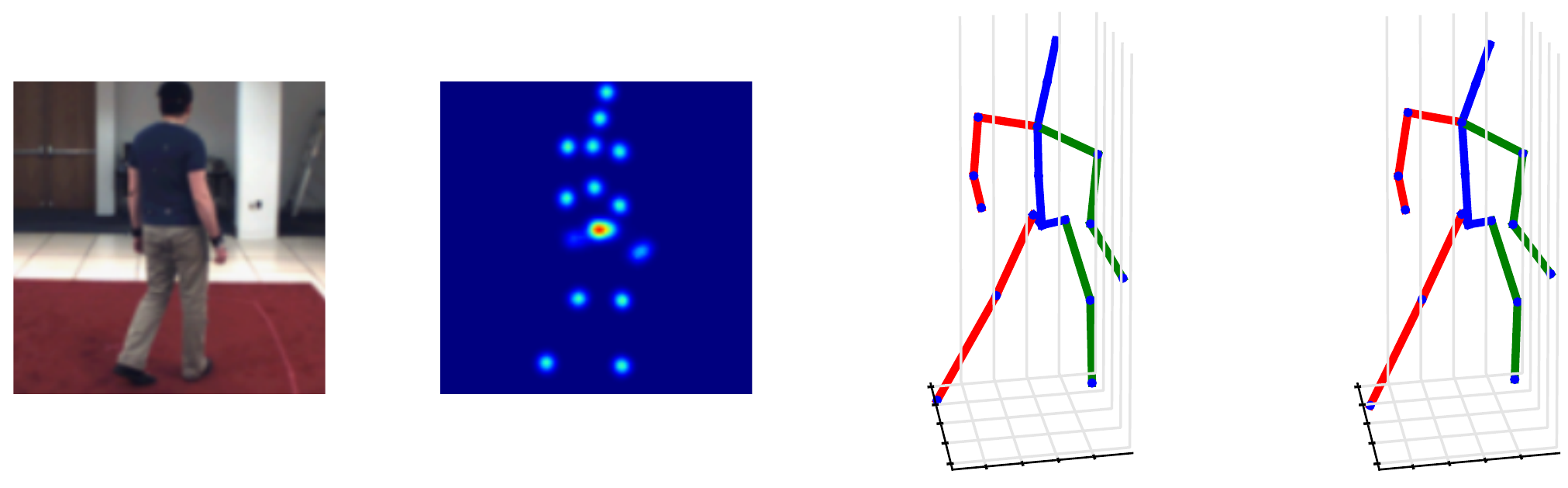} \\ \vspace{8mm}
			\includegraphics[width=0.45\linewidth, height=3.2cm]{he_6.pdf} \hspace{6mm}
			& \includegraphics[width=0.45\linewidth, height=3.2cm]{he_2.pdf} \\ \vspace{8mm}
			\includegraphics[width=0.45\linewidth, height=3.2cm]{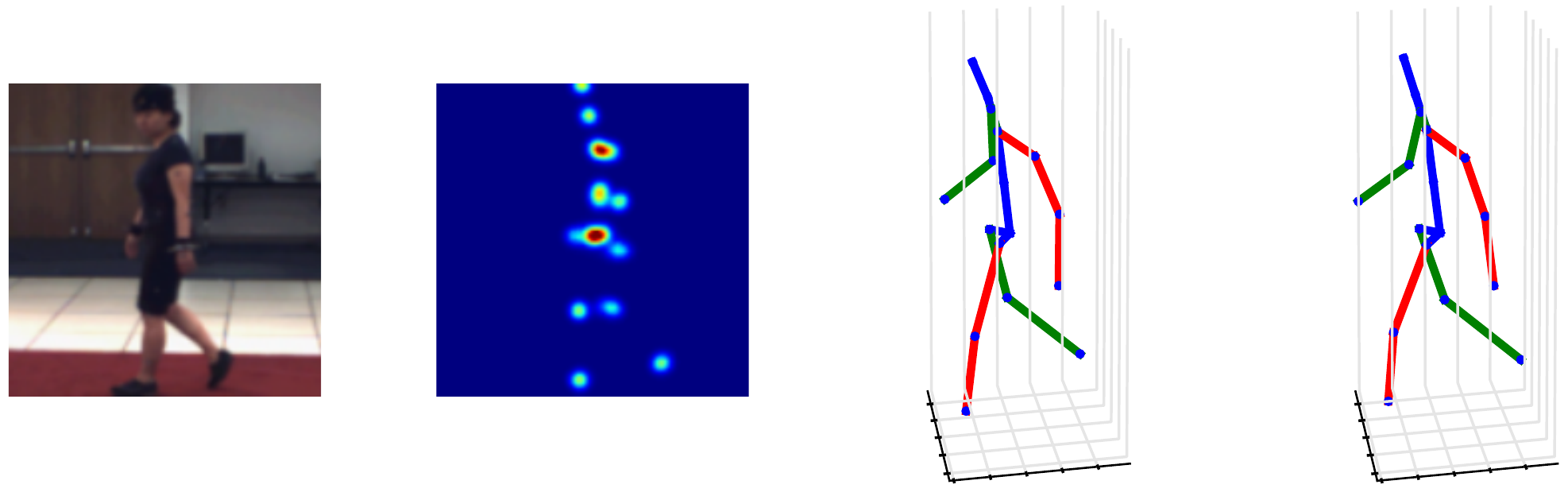} \hspace{6mm}
			& \includegraphics[width=0.45\linewidth, height=3.2cm]{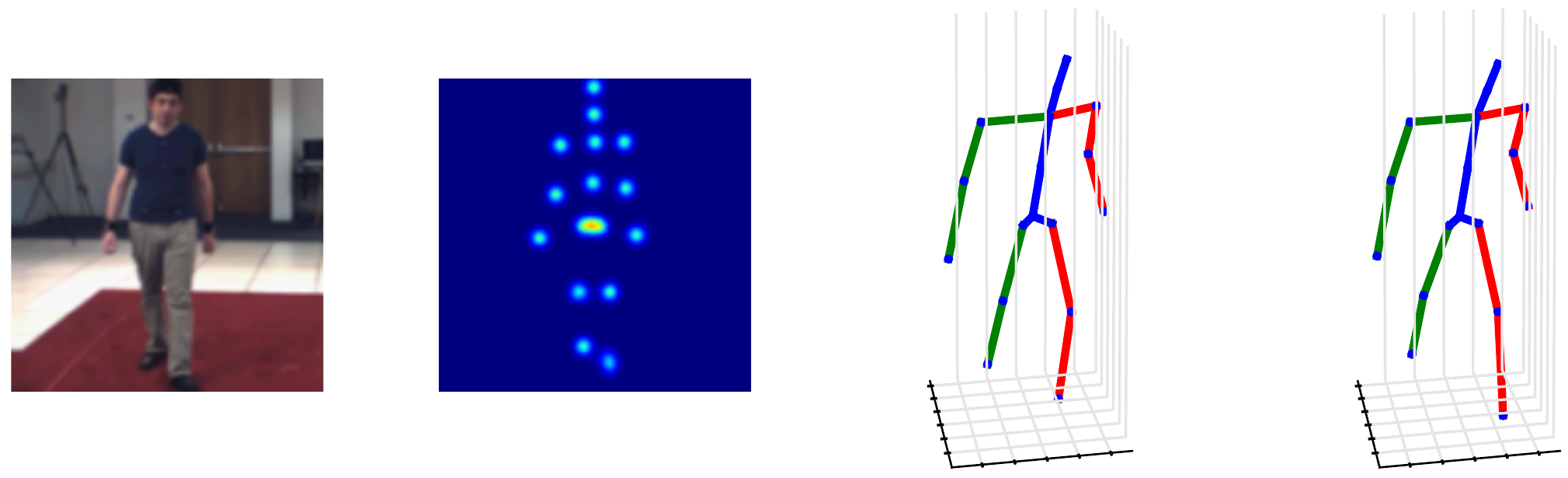} \\ \vspace{8mm}
			\includegraphics[width=0.45\linewidth, height=3.2cm]{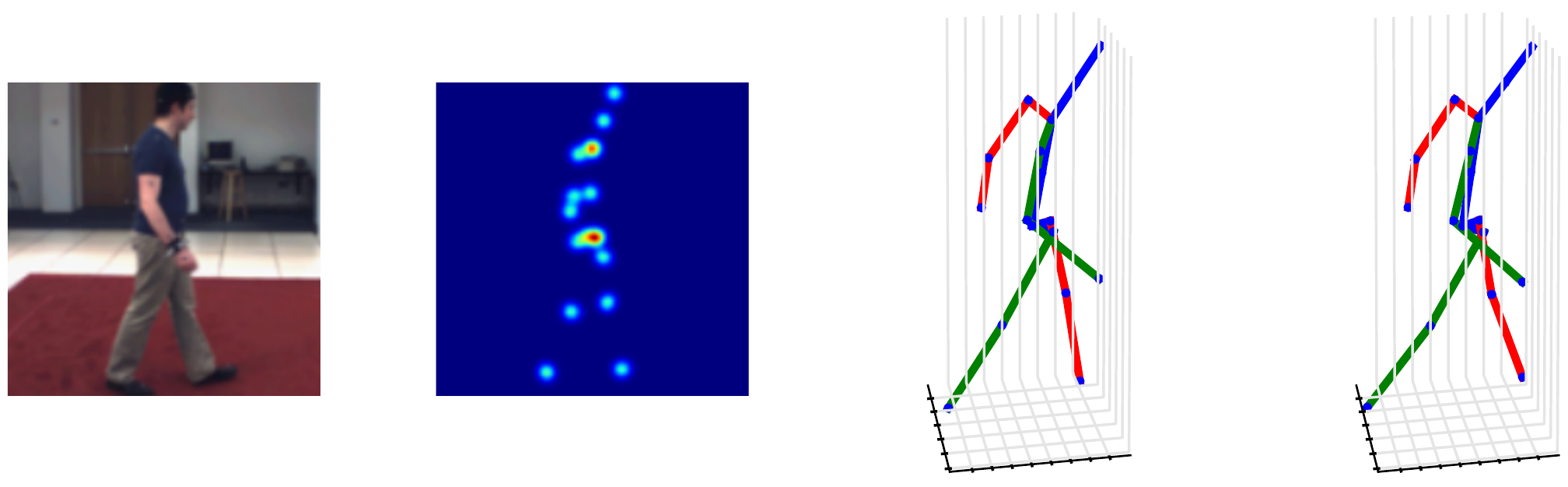} \hspace{6mm}
			& \includegraphics[width=0.45\linewidth, height=3.2cm]{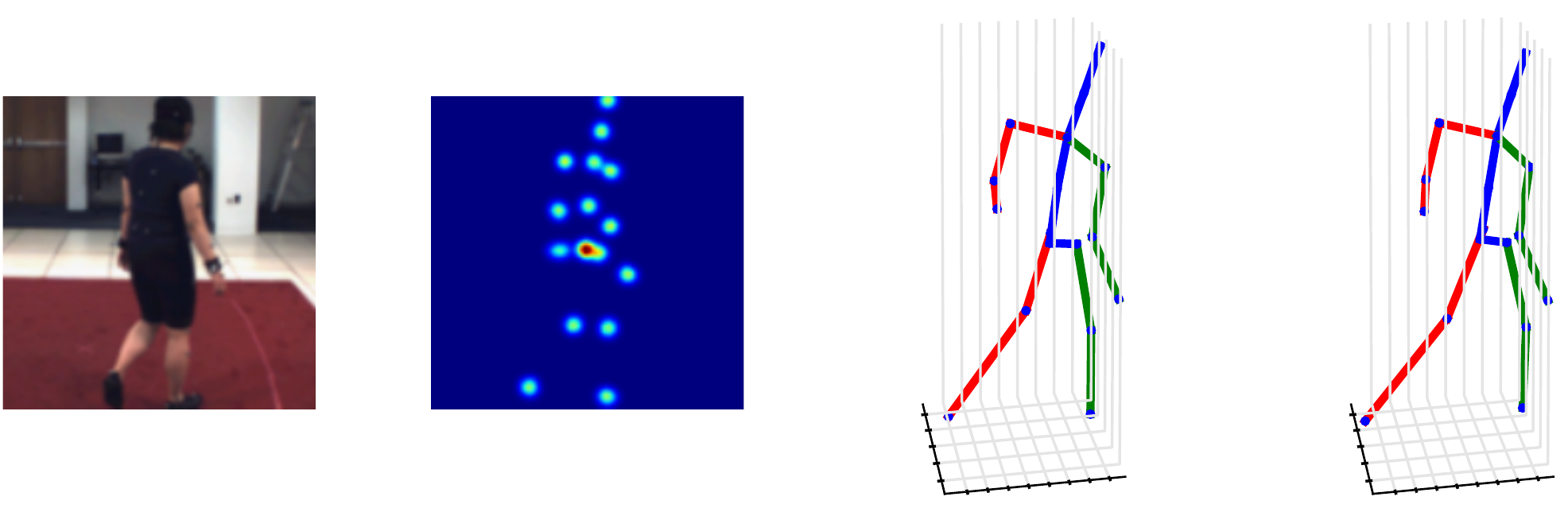} \\ \vspace{8mm}
			\includegraphics[width=0.45\linewidth, height=3.2cm]{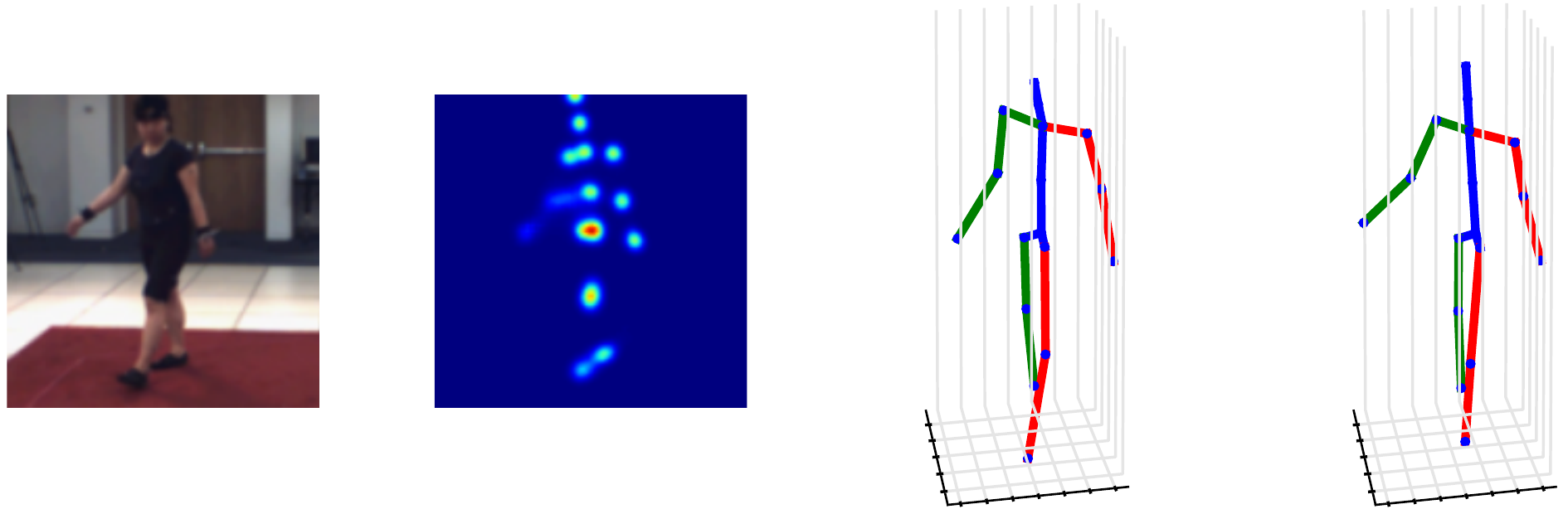} \hspace{6mm}
			& \includegraphics[width=0.45\linewidth, height=3.2cm]{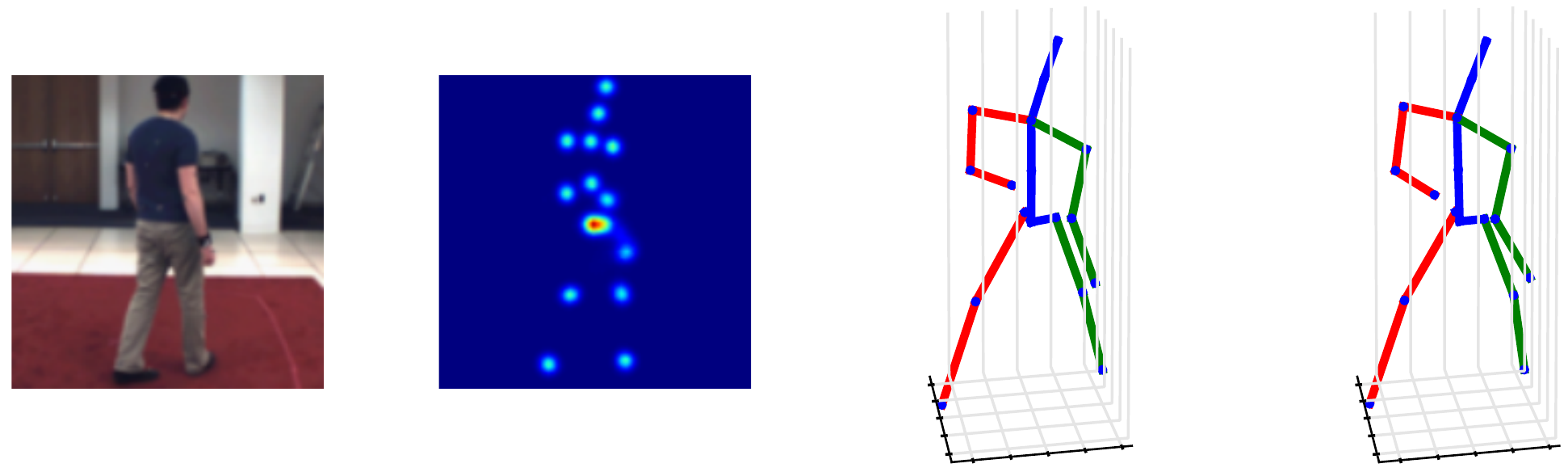} \\ \vspace{8mm}
			\footnotesize (a) Image \;\;\; (b) Confidence Map \;\;\; (c) Prediction \;\;\; (d) Ground-truth & \footnotesize \;\;\;\;\;\;\;\;\;\;\; (e) Image \;\;\; (f) Confidence Map \;\;\; (g) Prediction \;\;\; (h) Ground-truth \;\;\;
		\end{tabular}
	} \vspace{0mm}
	\caption{Pose  estimation  results  on  HumanEva-I. {\bf  (a, e)}  Input
		images. {\bf (b, f)} 2D joint  location confidence maps.
		{\bf (c, g)} Recovered pose. {\bf (d, h)} Ground truth. Best viewed in color.}
	\label{fig:results_he}
\end{figure*}

\begin{figure*}[tbph]
	\centering
	\scalebox{1}{
		\begin{tabular}{cc}
			\includegraphics[width=0.45\linewidth,height=1in]{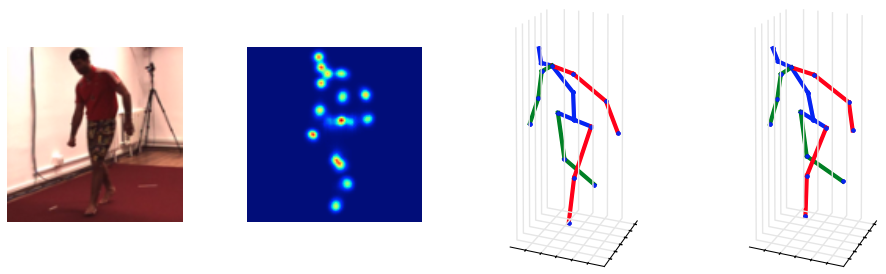} \hspace{6mm}
			& \includegraphics[width=0.45\linewidth]{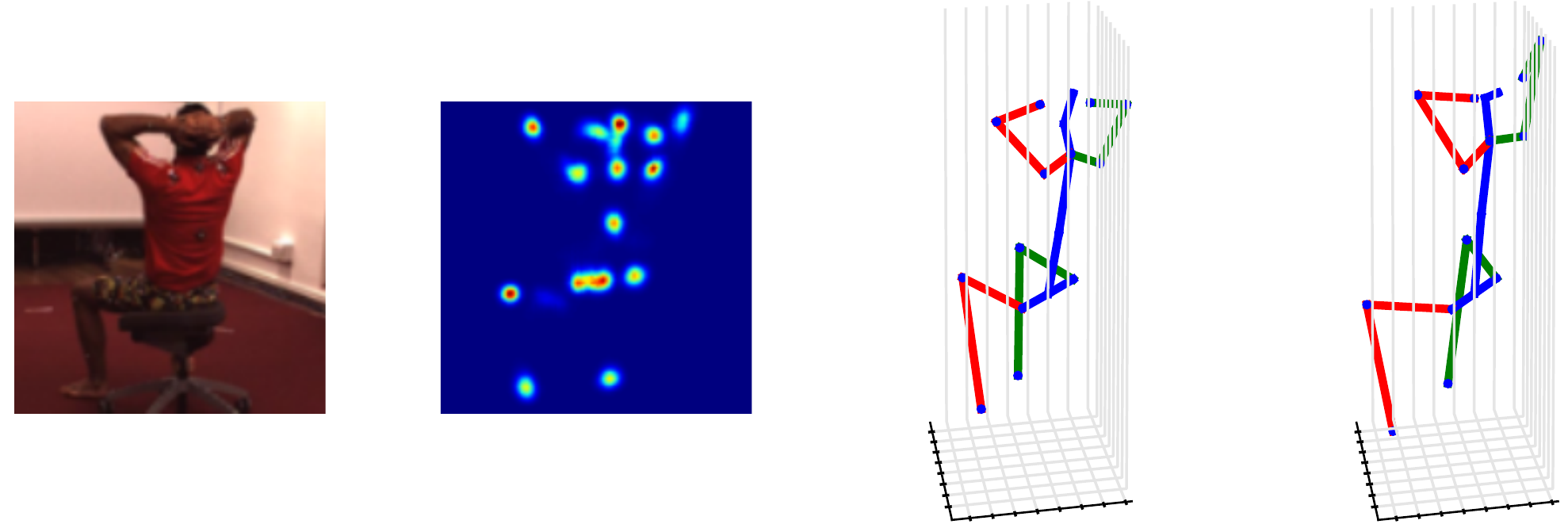} \\ \vspace{4mm}
			\includegraphics[width=0.45\linewidth,height=0.8in]{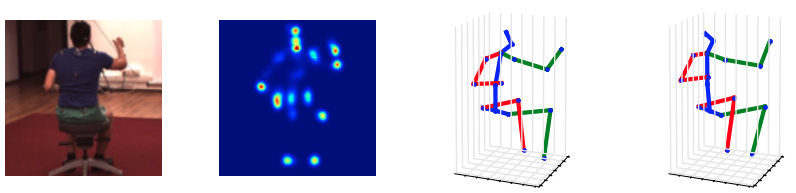} \hspace{6mm}		
			& \includegraphics[width=0.45\linewidth, height=1in]{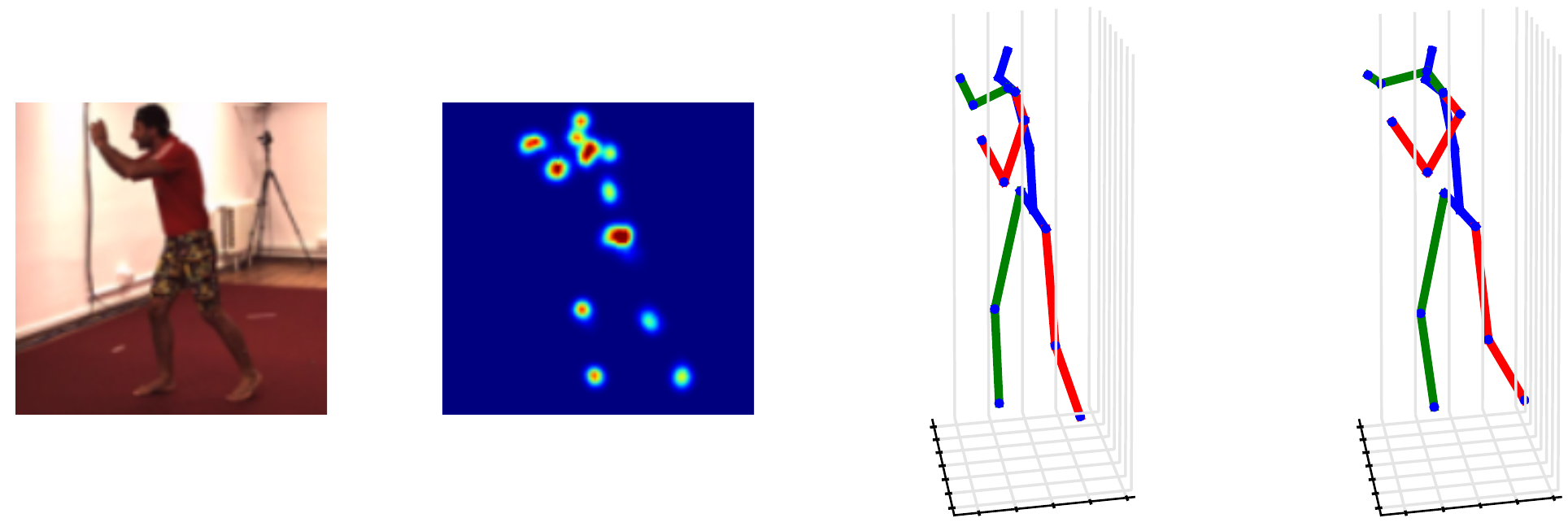} \\ \vspace{4mm}
			\includegraphics[width=0.45\linewidth,height=1in]{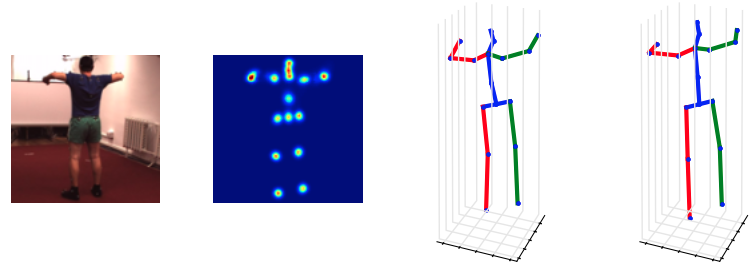} \hspace{6mm}
			& \includegraphics[width=0.45\linewidth]{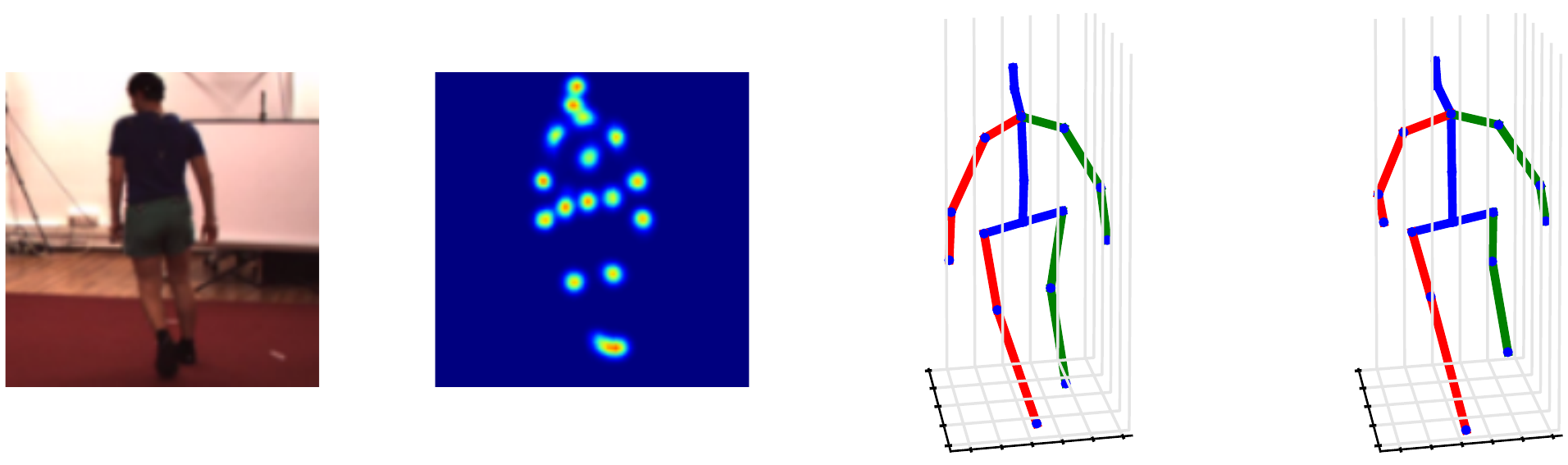} \\ \vspace{4mm}
			\includegraphics[width=0.45\linewidth,height=1in]{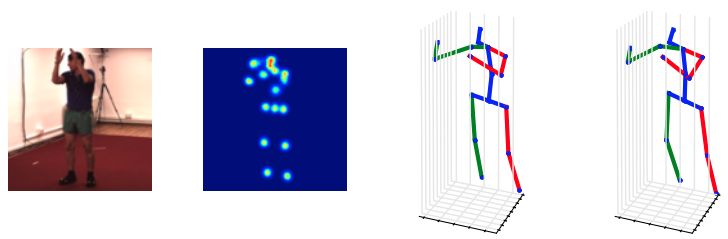} \hspace{6mm}
			& \includegraphics[width=0.45\linewidth]{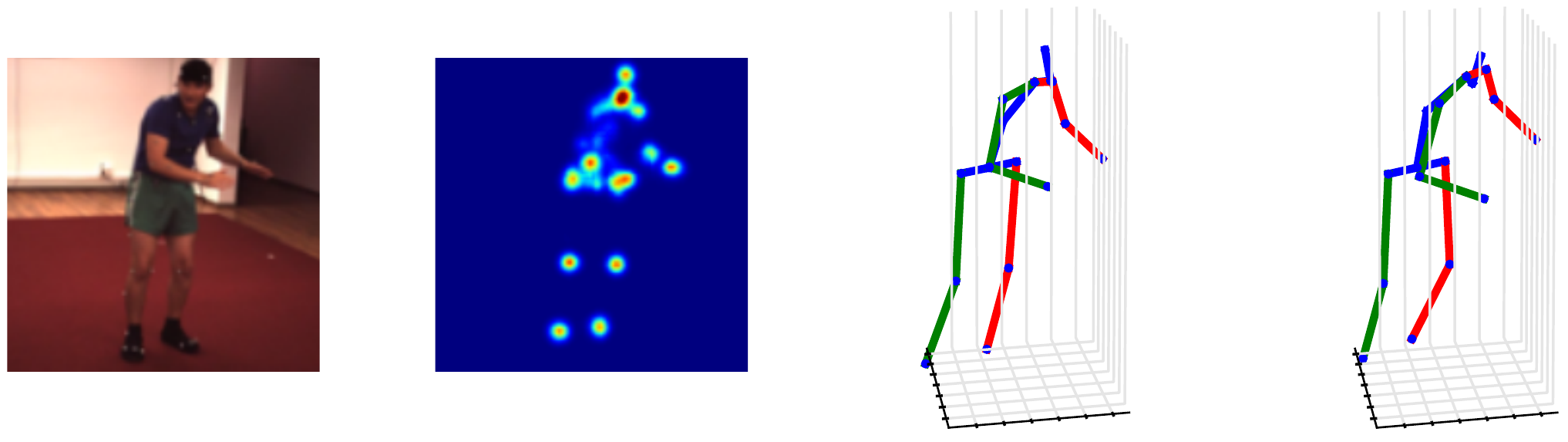} \\ \vspace{4mm}
			\footnotesize (a) Image \;\;\; (b) Confidence Map \;\;\; (c) Prediction \;\;\; (d) Ground-truth & \footnotesize \;\;\;\;\;\;\;\;\;\;\; (e) Image \;\;\; (f) Confidence Map \;\;\; (g) Prediction \;\;\; (h) Ground-truth \;\;\;
		\end{tabular}
	} \vspace{0mm}
	\caption{Pose  estimation  results  on  Human3.6m. {\bf  (a, e)}  Input
		images. {\bf (b, f)} 2D joint  location confidence maps.
		{\bf (c, g)} Recovered pose. {\bf (d, h)} Ground truth.
		Note that our method can  recover  the  3D  pose  in these  challenging  scenarios,  which
		involve significant amounts of self occlusion  and orientation ambiguity. Best viewed in color.}
	\label{fig:results_h36m}
\end{figure*}

\begin{figure*}[tbph]
	\centering
	\scalebox{1}{
		\begin{tabular}{cc}
			\includegraphics[width=0.45\linewidth]{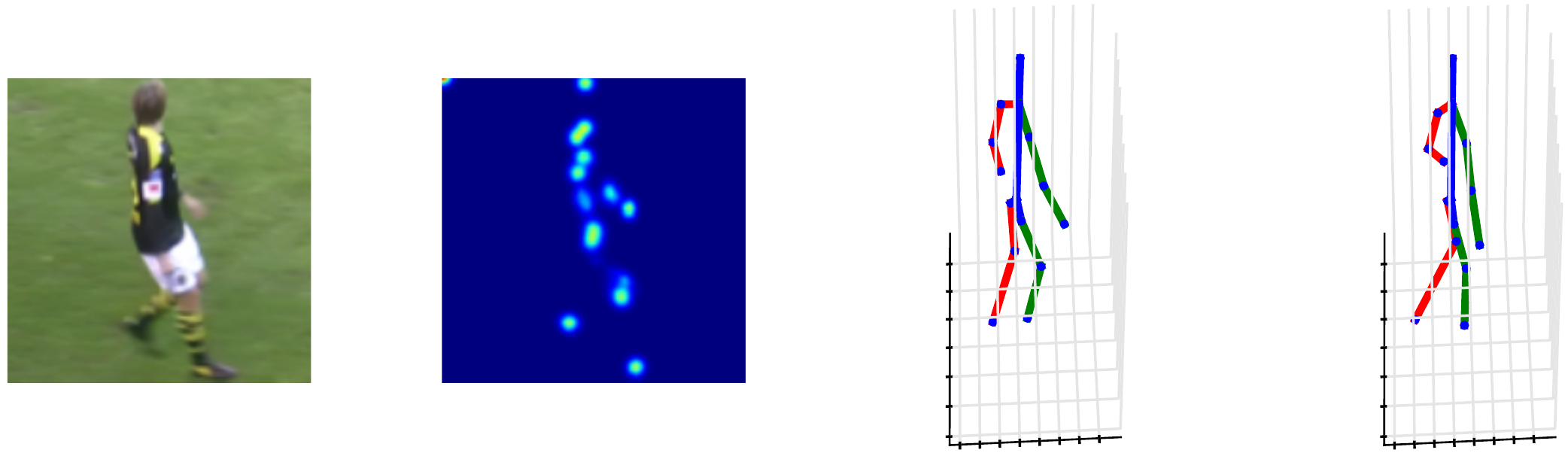} 
			&\includegraphics[width=0.45\linewidth]{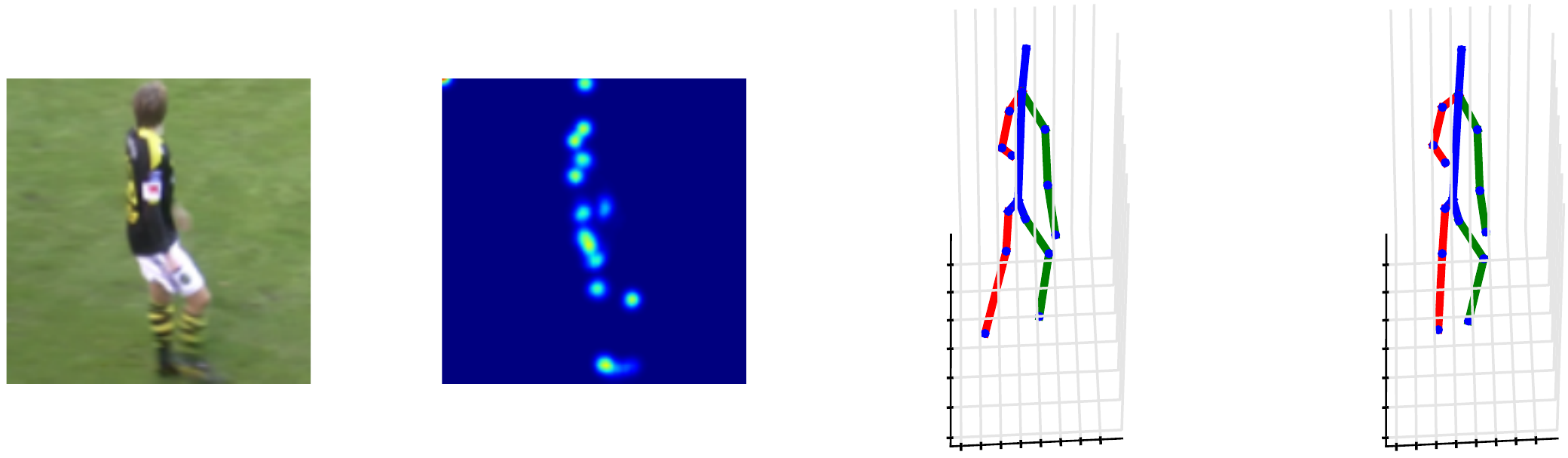} \\ \vspace{0mm}
			\includegraphics[width=0.45\linewidth]{kth_9.pdf} 
			&\includegraphics[width=0.45\linewidth]{kth_11.pdf} \\
			\footnotesize (a) Image \;\;\; (b) Confidence Map \;\;\; (c) Prediction \;\;\; (d) Ground-truth & \footnotesize \;\;\;\;\;\;\;\;\;\;\; (e) Image \;\;\; (f) Confidence Map \;\;\; (g) Prediction \;\;\; (h) Ground-truth \;\;\;
		\end{tabular}
	} \vspace{0mm}
	\caption{Pose  estimation  results  on  KTH Multiview Football II. {\bf  (a, e)}  Input
		images. {\bf (b, f)} 2D joint  location confidence maps.
		{\bf (c, g)} Recovered pose. {\bf (d, h)} Ground truth. Best viewed in color.}
	\label{fig:results_kth}
\end{figure*}

{\small
\bibliographystyle{ieee}
\bibliography{short,vision,learning}
}

\end{document}